\documentclass[letterpaper]{article} 
\usepackage[preprint]{aaai2027}
\usepackage[hyphens]{url}  
\usepackage{graphicx} 
\usepackage{natbib}  
\usepackage{caption} 
\usepackage{amsmath}
\usepackage{amssymb}
\usepackage{array}
\usepackage{placeins}
\usepackage{booktabs}
\usepackage{multirow}
\usepackage{tabularx}
\newcolumntype{Y}{>{\centering\arraybackslash}X}
\newcolumntype{G}{>{\centering\arraybackslash}p{0.14\columnwidth}}
\newcolumntype{L}{>{\raggedright\arraybackslash}X}
\newcommand{\methodname}{DynCur-Geo}
\newcommand{\ourmethod}{our method}
\newcommand{\sTenProtocolGraphic}{\includegraphics[width=0.88\textwidth]{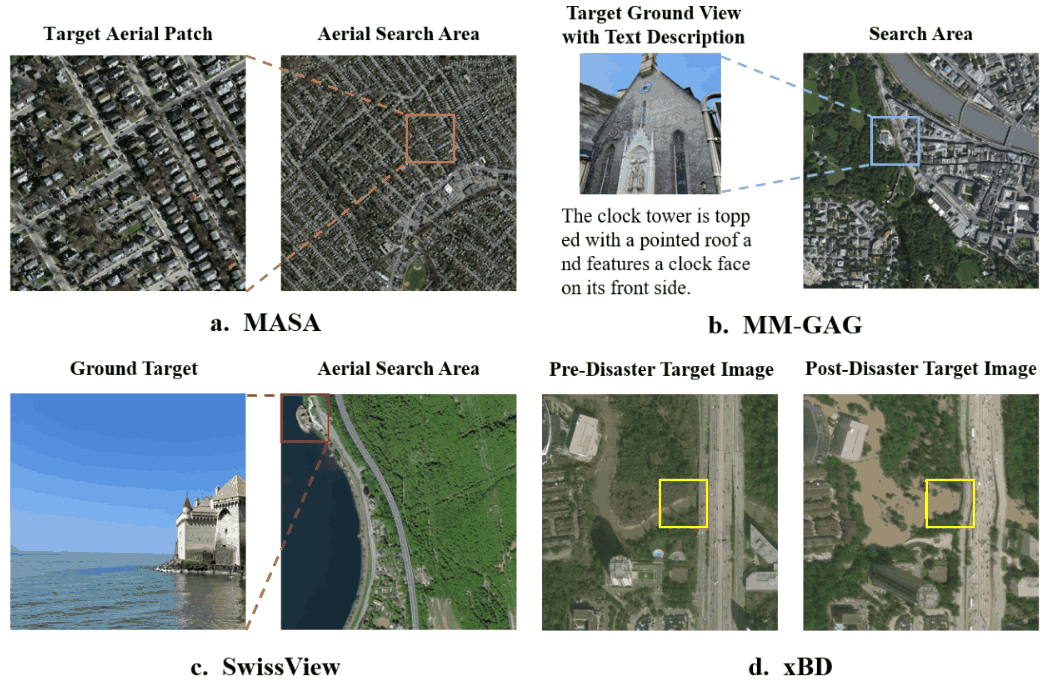}}
\newcommand{\suppwidetable}{\small\setlength{\tabcolsep}{3.0pt}\renewcommand{\arraystretch}{1.06}}
\newcommand{\suppnarrowtable}{\small\setlength{\tabcolsep}{4.0pt}\renewcommand{\arraystretch}{1.06}}

\makeatletter
\setcitestyle{numbers,square,comma}
\def\NAT@sort{\@ne}
\NAT@stdbstfalse
\makeatother

\title{{DynCur-Geo}: Dynamic Curiosity Reward Shaping for Multimodal Active Geo-Localization}
\author{Yiming Sun\textsuperscript{\rm 1},
Yang Zhang\textsuperscript{\rm 1},
Pengfei Zhu\textsuperscript{\rm 1}}
\affiliations{
    \textsuperscript{\rm 1}School of Automation, Southeast University, Nanjing 210096, China
}

\begin{document}

\maketitle

\begin{abstract}
Active geo-localization enables low-altitude UAVs to search for specified targets from limited local aerial observations, supporting time-sensitive applications such as search and rescue and emergency inspection. However, multimodal target cues, restricted views, and sparse feedback make it difficult to balance exploration with target convergence. Existing curiosity-driven methods assign a fixed intrinsic-reward weight throughout search, which can continue rewarding novelty after the agent nears the target and induce detours. We propose DynCur-Geo, a dynamic curiosity framework that adjusts prediction-error intrinsic reward according to remaining target distance. A distance-aware gate encourages early exploration and shifts the policy toward goal-directed behavior near the target, while potential-based reward shaping supplies dense progress guidance. Experiments across multimodal, cross-scene, disaster-affected, and long-range settings show consistent gains over active geo-localization baselines.
\end{abstract}

\section{Introduction}

Low-altitude UAVs can search large areas when direct access is difficult or rapid response is required. In search and rescue or emergency inspection, however, the target may be absent from the initial view, so the UAV must gather evidence, choose movements, and localize the target before the search budget is exhausted. This distinguishes active geo-localization (AGL) from conventional visual geo-localization, where a query is matched against a geotagged reference set or mapped to coordinates \cite{arandjelovic2016netvlad,hu2018cvmnet,berton2022benchmark,zheng2020university,zhu2021vigor,zhu2022transgeo,ji2025game4loc}. AGL instead poses localization as sequential decisions over local aerial observations, making an efficient learned policy essential \cite{pirinen2023airloc,sarkar2024gomaa,mi2025geoexplorer}.

\begin{figure}[t]
\centering
\includegraphics[width=\columnwidth]{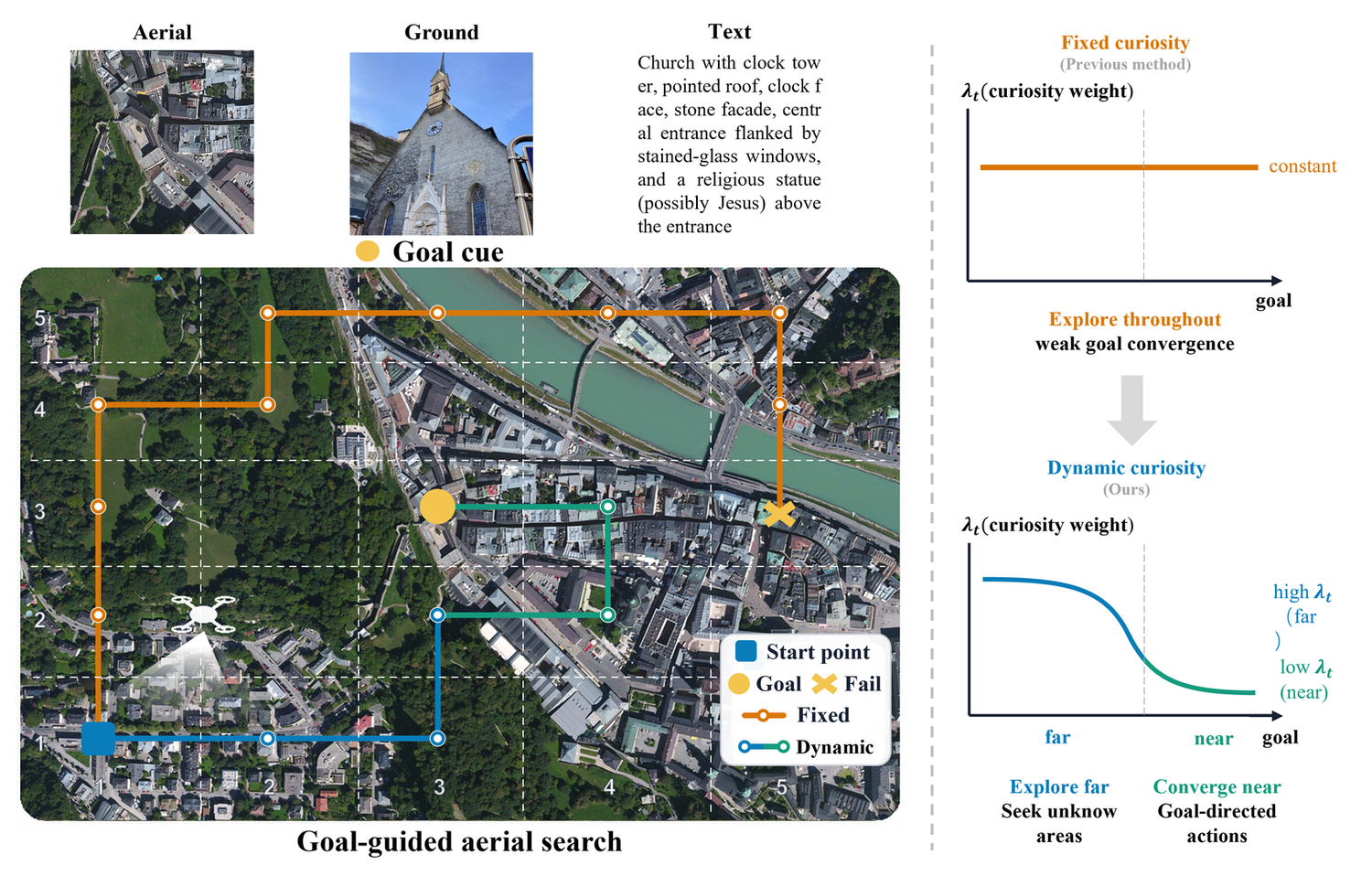}
\caption{Motivation for distance-gated curiosity. GeoExplorer uses a fixed curiosity weight that can weaken late-stage convergence by rewarding novelty throughout search (orange). Our dynamic gate weights intrinsic reward more strongly far from the target and less near it, promoting exploration followed by goal-directed convergence (blue--green).}
\label{fig:distance_gated_motivation}
\end{figure}

Targets may be specified by a previous aerial crop, a ground-view photograph, or a short text description. GOMAA-Geo~\cite{sarkar2024gomaa} extends AGL to this goal-modality-agnostic setting, where the policy must align these cues with local aerial observations. The cue and local views may differ strongly in appearance and viewpoint, directional evidence can remain ambiguous for several steps, and every exploratory action consumes budget.

Curiosity-driven exploration helps under sparse feedback. GeoExplorer~\cite{mi2025geoexplorer} uses action-state prediction error as an intrinsic reward and combines it with the external reward using a fixed curiosity weight throughout search. Although this encourages informative transitions, prediction error measures novelty rather than goal relevance, so it can preserve exploratory pressure after the policy enters the target neighborhood, where convergence is more valuable than unfamiliar states.

To address this limitation, we propose DynCur-Geo, a dynamic curiosity framework for multimodal AGL. Fig.~\ref{fig:distance_gated_motivation} contrasts the fixed curiosity weight of GeoExplorer~\cite{mi2025geoexplorer} with our trajectory-dependent design. Broad exploration is useful early, whereas later decisions should prioritize stable progress. We implement this transition with distance-gated curiosity reward shaping: intrinsic reward is weighted more far from the target and downweighted as the policy approaches it. We also add potential-based reward shaping (PBRS)~\cite{ng1999policy,adamczyk2025bsrs} to provide dense target-progress feedback and reinforce goal-directed convergence.

Our main contributions are summarized as follows:
\begin{itemize}
    \item We introduce DynCur-Geo, a dynamic-curiosity framework for multimodal active geo-localization that uses multimodal target cues and observation features within one sequential search policy.
    \item We develop a distance-aware curiosity gate that adapts prediction-error intrinsic reward along the trajectory, supporting broad exploration far from the target and precise convergence near it.
    \item We integrate potential-based target-progress shaping with dynamic curiosity and demonstrate consistent gains over existing methods across multimodal, cross-scene, disaster-affected, and long-range search settings.
\end{itemize}

\section{Related Work}

\paragraph{Visual and Multimodal Geo-Localization.}
Visual geo-localization emphasizes representation learning for retrieval or coordinate estimation. Place-recognition models learn descriptors for large reference databases \cite{arandjelovic2016netvlad,berton2022benchmark}, while cross-view methods match images from ground cameras, drones, and satellites \cite{workman2015wide,vo2016street,hu2018cvmnet,liu2019oricnn,shi2019safa,yan2022crossloc,zhu2021vigor,zhu2022transgeo}. Recent work considers limited fields of view, orientation uncertainty, drone-view queries, object-level targets, broader geographic coverage, and multimodal queries \cite{zheng2020university,ji2025game4loc,zheng2026geoxbench,radford2021clip,cepeda2023geoclip,ji2025mmgeo,zhang2026mopuav}. Contrastive and vision-language models further connect visual observations with geographic or textual representations \cite{radford2021clip,cepeda2023geoclip}, and multimodal UAV localization extends queries with text, prompts, depth, point clouds, or multiple image views \cite{ji2025mmgeo,zhang2026mopuav}.

These works motivate multimodal localization representations, but their protocols differ from AGL. Retrieval methods compare a query against a reference database or map; AGL observes only the current aerial patch, chooses the next move before seeing future patches, and is judged by whether it reaches the target cell within budget. Thus, the direct baselines are active-search policies under the same movement protocol.

\paragraph{Active Geo-Localization and Aerial Navigation.}
AiRLoc~\cite{pirinen2023airloc} abstracts search-and-rescue-style aerial search as a reinforcement learning problem in which a policy localizes an aerial-view target from partial glimpses. GOMAA-Geo~\cite{sarkar2024gomaa} extends the formulation to multimodal goal cues and combines cross-modal alignment, supervised pretraining, and actor-critic policy learning for aerial, ground-view, and text goals. GeoExplorer~\cite{mi2025geoexplorer} adds action-state dynamics modeling and next-state prediction error as intrinsic reward. These closest AGL baselines use the same controlled grid-based setting; our work studies how prediction-error curiosity should be weighted as the policy moves from broad exploration to target convergence.

Related aerial and embodied navigation tasks study language instruction following, continuous motion, or reaching visible goals \cite{anderson2018vision,liu2023aerialvln,zhu2017targetdriven}. In contrast, AGL uses grid-based target localization, with performance measured by success rate and final grid distance under a fixed movement budget.

\paragraph{Curiosity and Reward Shaping.}
Intrinsic motivation methods use prediction error, novelty, reachability, or uncertainty to encourage exploration when external rewards are sparse or delayed \cite{schmidhuber1991curiosity,stadie2015incentivizing,bellemare2016unifying,pathak2017curiosity,burda2018large,burda2019rnd,savinov2019episodic,sekar2020plan2explore}. In visual reinforcement learning, prediction-error curiosity learns a dynamics model and rewards hard-to-predict transitions. This fits AGL's repetitive local aerial patches and late target visibility, where sparse success reward gives little early feedback.

However, curiosity is not inherently task-aligned: exploration should increase the chance of reaching the target, not merely visit surprising states. PBRS~\cite{ng1999policy,adamczyk2025bsrs} adds dense progress feedback through a potential function while preserving policy invariance under standard assumptions. Our method uses prediction-error curiosity for exploration and combines a distance gate with PBRS to tie it to target convergence.

\begin{figure*}[t]
\centering
\includegraphics[width=0.92\textwidth]{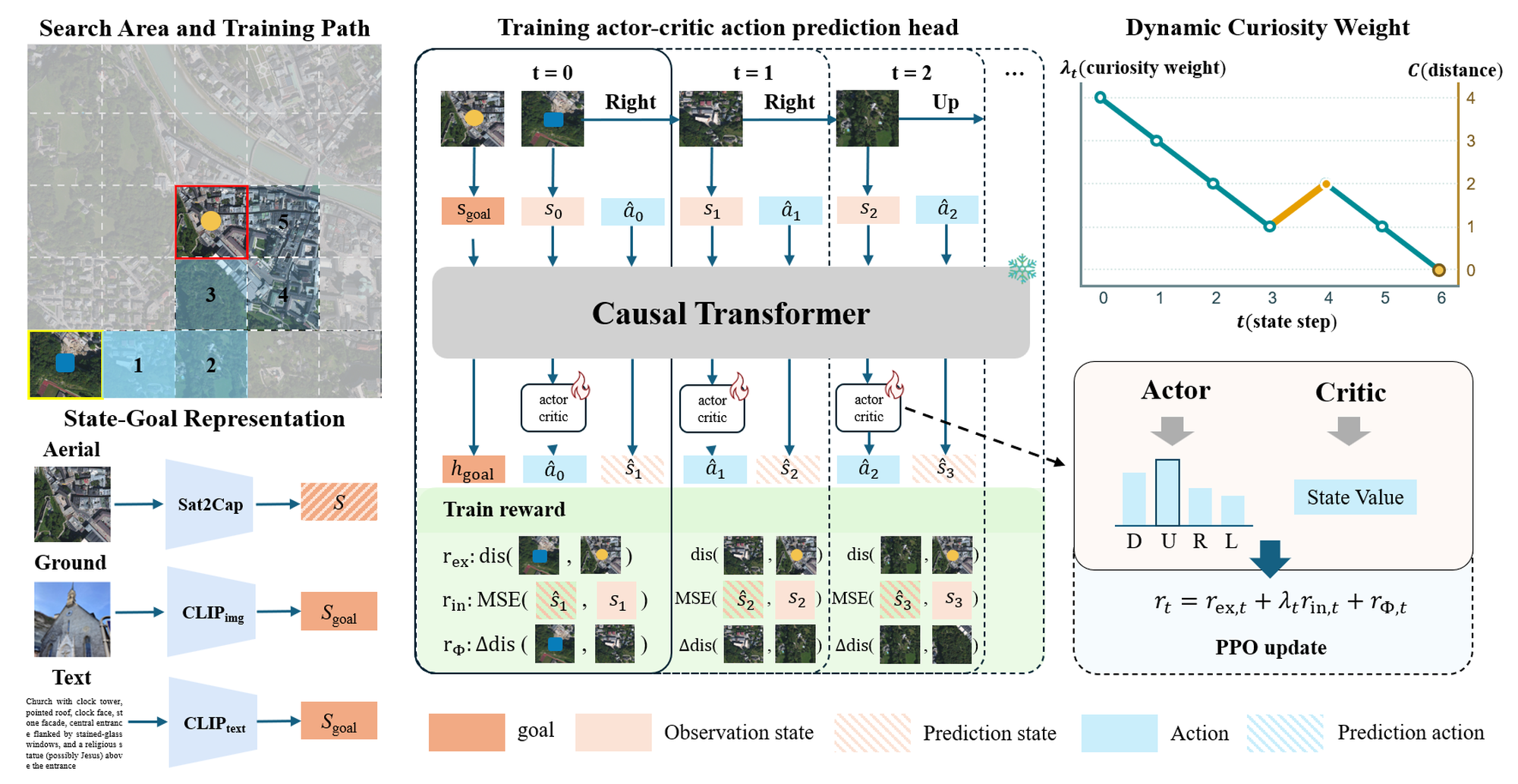}
\caption{Training pipeline and reward construction. Target cues and local aerial observations form a trajectory history processed by a frozen GeoExplorer dynamics model. Prediction error provides intrinsic reward, whose weight is reduced near the target and combined with external progress and PBRS rewards for PPO training.}
\label{fig:method_overview}
\end{figure*}

\section{Method}

\paragraph{Problem Setup.}
We formulate multimodal active geo-localization as a goal-conditioned sequential decision problem on a discretized aerial region. The region is divided into \(K\times K\) cells, and each cell \(p\) corresponds to an aerial image patch \(I_p\). For each search episode, the policy first receives a target cue \(g\), which may be an aerial image, a ground-view image, or a text description. The cue is associated with a target cell \(p_g\), but the target location is hidden during inference. Starting from cell \(p_0\), the policy observes only the aerial patch at its current cell \(p_t\), selects an action \(a_t\) from \(\mathcal{A}=\{\mathrm{up},\mathrm{right},\mathrm{down},\mathrm{left}\}\), and moves to the next cell if the action is legal. A search episode is successful if the policy reaches \(p_g\) within the budget \(B\).

At each episode start, the policy receives only the target cue and the observation at \(p_0\); after each action, it receives the resulting local observation. The target cell is used only during training to construct dynamics targets and rewards. During inference, the policy does not evaluate target distance, reward terms, or an oracle planner; it acts from the learned representation, history, and legal-action mask. The resulting training pipeline and reward construction are summarized in Fig.~\ref{fig:method_overview}.

\paragraph{Multimodal Feature Representation.}
Let \(m\) denote the modality of the target cue \(g\). Following the feature representation used in multimodal AGL \cite{sarkar2024gomaa,mi2025geoexplorer}, Sat2Cap~\cite{dhakal2024sat2cap} encodes aerial images for both aerial target cues and search observations, while the CLIP ViT-B/32~\cite{radford2021clip} image and text encoders represent ground-view and text targets. We denote the corresponding frozen modality encoder by \(E_m\), which maps the target cue to a shared 512-dimensional target embedding:
\begin{equation}
q = E_m(g), \quad m\in\{\mathrm{aerial},\mathrm{ground},\mathrm{text}\}.
\label{eq:target_embedding}
\end{equation}
For search observations, the same aerial encoder \(E_a\) maps the current patch \(I_{p_t}\) to an observation feature \(v_t=E_a(I_{p_t})\). We augment this feature with a lightweight sinusoidal grid-position encoding and denote the resulting location-aware feature as \(\tilde v_t\). Given a learned action embedding \(e(a_i)\), the target, observation, and action features up to step \(t\) form the causal input sequence
\begin{equation}
Z_t=[q,\tilde v_0,e(a_0),\tilde v_1,\ldots,e(a_{t-1}),\tilde v_t].
\label{eq:history_sequence}
\end{equation}
We reuse the pretrained causal Transformer \(F_\theta\) from GeoExplorer~\cite{mi2025geoexplorer} as the sequence model. With input projections from the 512-dimensional feature space and learned action tokens, it maps the causal sequence to a state representation
\begin{equation}
h_t=F_\theta(Z_t),
\label{eq:history_encoder}
\end{equation}
where \(\theta\) denotes the model parameters. The policy therefore combines multimodal target and observation features with the action sequence instead of relying on the current patch alone.

\paragraph{Action-State Dynamics Modeling.}
The adopted GeoExplorer~\cite{mi2025geoexplorer} action-state model was pretrained on generated trajectories. At each step, it predicts two targets: a multi-label action vector \(y_t\), where each entry indicates whether the corresponding movement reduces the grid distance to the goal, and the next-state feature after the action-conditioned transition. The dynamics-modeling loss is
\begin{equation}
\mathcal{L}_{\mathrm{DM}}=
\mathcal{L}_{\mathrm{act}}+\alpha\mathcal{L}_{\mathrm{state}},
\label{eq:dynamics_loss}
\end{equation}
where \(\mathcal{L}_{\mathrm{act}}\) is the action prediction loss, \(\mathcal{L}_{\mathrm{state}}\) is the state-feature prediction loss, and \(\alpha\) balances the two terms. We use binary cross-entropy for \(\mathcal{L}_{\mathrm{act}}\) and mean-squared error for \(\mathcal{L}_{\mathrm{state}}\). This objective trains the sequence model to encode both goal-directed action preferences and action-conditioned changes in aerial observations.

During PPO~\cite{schulman2017ppo} training, the pretrained sequence model is frozen and used as a stable history encoder and dynamics model. Let \(\hat v_{t+1}\) denote the predicted next-state feature and \(\tilde v_{t+1}\) denote the observed next-state feature after executing \(a_t\). The prediction error is
\begin{equation}
e_t=\left\|\hat v_{t+1}-\tilde v_{t+1}\right\|_2^2.
\label{eq:prediction_error}
\end{equation}
This error is the transition novelty signal used for intrinsic reward.

\paragraph{Distance-Gated External and Intrinsic Rewards.}
The reinforcement-learning stage follows the distance-based external reward commonly used in AGL, which gives dense feedback according to whether an action moves the policy toward the target. We use this reward as the task-progress signal and write it as
\begin{equation}
r_{\mathrm{ex},t} =
\begin{cases}
r_g & p_{t+1}=p_g\\
r_p & \Delta(p_{t+1},p_g)<b_t\\
r_n & \text{otherwise}
\end{cases},
\label{eq:external_reward}
\end{equation}
where \(p_{t+1}\) is the cell after action \(a_t\), \(r_g\) is the goal-reaching reward, \(r_p\) is the progress reward, and \(r_n<0\) penalizes non-progress moves. The function \(\Delta(p,p_g)\) measures distance-based progress between cell \(p\) and target cell \(p_g\), while \(b_t\) records the best accepted progress score along the current trajectory. This term follows the prior dense AGL reward design and uses the known target location only during training.

The intrinsic reward is derived from the frozen dynamics model:
\begin{equation}
r_{\mathrm{in},t}=\psi(e_t),
\label{eq:intrinsic_reward}
\end{equation}
where \(\psi(\cdot)\) normalizes prediction errors to a scale comparable with the external reward. The resulting signal gives dense feedback when target progress is hard to infer from local observations, but it is goal-agnostic and therefore does not distinguish target-helpful novelty from target-irrelevant novelty. We therefore gate curiosity by the remaining target distance. Let \(d_1(p,p_g)\) be the Manhattan distance between cell \(p\) and the target cell. We first define a normalized distance ratio
\begin{equation}
\rho_t=\mathrm{clip}\left(
\frac{d_1(p_t,p_g)}
{\max(d_1(p_0,p_g),1)},0,1
\right),
\label{eq:distance_ratio}
\end{equation}
where \(p_t\) is the current cell, \(p_0\) is the start cell, and the denominator is lower bounded by one for numerical stability. The distance ratio then defines a family of curiosity gates,
\begin{equation}
\lambda_t=\lambda_{\min}+(1-\lambda_{\min})f(\rho_t),
\label{eq:distance_gate}
\end{equation}
where \(\lambda_{\min}\) is the lower bound of the curiosity weight and \(f(\rho_t)\) controls the distance schedule. Eq.~\eqref{eq:distance_gate} assigns a larger curiosity weight far from the target and reduces that weight near the target. The final model uses a linear gate, \(f(\rho_t)=\rho_t\). The ablation study also compares a constant gate \(f(\rho_t)=0\), a sine gate \(f(\rho_t)=\sin(\pi\rho_t/2)\), and a power gate \(f(\rho_t)=\rho_t^2\). The special case \(\lambda_t=1\) denotes ungated curiosity.

\paragraph{Potential-Based Reward Shaping.}
The second reward component is potential-based reward shaping (PBRS), which supplies smooth distance feedback without changing the inference-time inputs. For a \(K\times K\) grid, the maximum Manhattan distance is \(d_{\max}=2(K-1)\). We define the potential function
\begin{equation}
\Phi(p)=-\frac{d_1(p,p_g)}{d_{\max}},
\label{eq:potential}
\end{equation}
and the shaping reward
\begin{equation}
r_{\Phi,t}=\beta\left(\gamma\Phi(p_{t+1})-\Phi(p_t)\right),
\label{eq:pbrs_reward}
\end{equation}
where \(\beta\) controls the strength of shaping and \(\gamma\) is the discount factor. The final training reward is
\begin{equation}
r_t=r_{\mathrm{ex},t}+\lambda_t r_{\mathrm{in},t}+r_{\Phi,t}.
\label{eq:training_reward}
\end{equation}
This wraps the reward design into three training-only signals: the prior distance-based external reward supplies task feedback, the gated intrinsic reward encourages exploration when target evidence is still sparse, and PBRS supplies dense target-progress shaping. At inference time, none of these reward terms or target-distance quantities is evaluated.

\paragraph{Policy Learning and Inference.}
With the frozen history encoder \(F_\theta\), we train an actor-critic policy using PPO. The actor outputs a distribution \(\pi(a|h_t)\) over actions \(a\in\mathcal{A}\), and the critic estimates the value of \(h_t\). Let \(M_t(a)\in\{0,1\}\) be the legal-action mask at step \(t\), where invalid boundary actions have value zero. The masked policy is
\begin{equation}
\bar{\pi}(a|h_t)=
\frac{M_t(a)\pi(a|h_t)}
{\sum_{a'\in\mathcal{A}}M_t(a')\pi(a'|h_t)}.
\label{eq:masked_policy}
\end{equation}
During inference, model parameters are fixed and the reward terms are not evaluated. The policy updates its history after each observation and selects the highest-probability legal action,
\begin{equation}
a_t=\arg\max_{a\in\mathcal{A},M_t(a)=1}\bar{\pi}(a|h_t),
\label{eq:greedy_action}
\end{equation}
until it reaches the target cell or exhausts the search budget. Thus, the reported gains come from the policy learned under the training reward rather than from test-time access to target distance.

\section{Experiments}

\begin{table*}[!t]
\centering
\begin{minipage}[t]{\columnwidth}
\centering
{\small
\setlength{\tabcolsep}{2.4pt}
\begin{tabularx}{\linewidth}{@{}c@{\hspace{0.55em}}l@{\hspace{0.25em}}*{6}{Y@{}}}
\toprule
& Method & \(C=4\) & \(C=5\) & \(C=6\) & \(C=7\) & \(C=8\) & Avg. \\
\midrule
\multirow{7}{*}{A} & Random & 0.1447 & 0.0936 & 0.0553 & 0.0170 & 0.0298 & 0.0681 \\
& PPO & 0.1745 & 0.1447 & 0.1574 & 0.2468 & 0.2723 & 0.1991 \\
& DiT & 0.1660 & 0.2255 & 0.2766 & 0.2681 & 0.1915 & 0.2255 \\
& AiRLoc & \textbf{0.5660} & \textbf{0.4894} & 0.4894 & 0.4383 & 0.3191 & 0.4604 \\
& GOMAA-Geo & 0.3574 & 0.3872 & 0.5532 & 0.6298 & 0.7404 & 0.5336 \\
& GeoExplorer & 0.3489 & 0.3745 & 0.5574 & 0.6851 & 0.6766 & 0.5285 \\
& \textbf{DynCur-Geo} & 0.2681 & 0.3702 & \textbf{0.6809} & \textbf{0.8426} & \textbf{0.9234} & \textbf{0.6170} \\
\midrule
\multirow{3}{*}{G} & GOMAA-Geo & 0.3489 & 0.3745 & 0.5617 & 0.7064 & 0.7702 & 0.5523 \\
& GeoExplorer & \textbf{0.4043} & 0.3745 & 0.5319 & 0.6681 & 0.7149 & 0.5387 \\
& \textbf{DynCur-Geo} & 0.2766 & \textbf{0.4468} & \textbf{0.6511} & \textbf{0.9106} & \textbf{0.9106} & \textbf{0.6391} \\
\midrule
\multirow{3}{*}{T} & GOMAA-Geo & 0.3447 & 0.4213 & 0.5745 & 0.6596 & 0.7362 & 0.5472 \\
& GeoExplorer & \textbf{0.3957} & 0.3660 & 0.4553 & 0.5660 & 0.6085 & 0.4783 \\
& \textbf{DynCur-Geo} & 0.2681 & \textbf{0.4298} & \textbf{0.6596} & \textbf{0.8511} & \textbf{0.9149} & \textbf{0.6247} \\
\bottomrule
\end{tabularx}
\par}
\captionof{table}{Success rate (SR) on MM-GAG~\cite{sarkar2024gomaa} under \(5\times5\), \(B=10\), and \(C=4,\ldots,8\). A/G/T denote aerial-image, ground-view, and text targets. PPO, DiT, and AiRLoc are evaluated only for aerial-image targets.}
\label{tab:mmgag_sr}
\end{minipage}
\hfill
\begin{minipage}[t]{\columnwidth}
\centering
{\small
\setlength{\tabcolsep}{2.4pt}
\begin{tabularx}{\linewidth}{@{}c@{\hspace{0.55em}}l@{\hspace{0.25em}}*{6}{Y@{}}}
\toprule
& Method & \(C=4\) & \(C=5\) & \(C=6\) & \(C=7\) & \(C=8\) & Avg. \\
\midrule
\multirow{7}{*}{A} & Random & 3.1404 & 3.5617 & 3.8468 & 4.3362 & 4.9021 & 3.9574 \\
& PPO & 3.5331 & 3.5684 & 3.8095 & 3.9232 & 4.5119 & 3.6545 \\
& DiT & 3.2596 & 3.3021 & 3.0468 & 3.3915 & 4.4766 & 3.4953 \\
& AiRLoc & \textbf{1.0979} & \textbf{1.2979} & 1.3064 & 1.5362 & 2.0723 & 1.4621 \\
& GOMAA-Geo & 2.4340 & 1.9574 & 1.3106 & 0.9660 & 0.6553 & 1.4647 \\
& GeoExplorer & 2.2213 & 1.9702 & 1.3277 & 0.9021 & 0.8681 & 1.4579 \\
& \textbf{DynCur-Geo} & 2.5447 & 1.8553 & \textbf{0.8426} & \textbf{0.3532} & \textbf{0.2383} & \textbf{1.1668} \\
\midrule
\multirow{3}{*}{G} & GOMAA-Geo & 2.3830 & 1.8340 & 1.2170 & 0.6596 & 0.6213 & 1.3430 \\
& GeoExplorer & \textbf{2.0170} & 1.9106 & 1.3277 & 0.8596 & 0.8936 & 1.4017 \\
& \textbf{DynCur-Geo} & 2.4766 & \textbf{1.7447} & \textbf{0.8681} & \textbf{0.2000} & \textbf{0.2383} & \textbf{1.1055} \\
\midrule
\multirow{3}{*}{T} & GOMAA-Geo & 2.3064 & 1.8298 & 1.3021 & 0.7915 & 0.7404 & 1.3940 \\
& GeoExplorer & \textbf{2.1191} & 2.0383 & 1.5915 & 1.2596 & 1.2851 & 1.6587 \\
& \textbf{DynCur-Geo} & 2.5106 & \textbf{1.6085} & \textbf{0.9191} & \textbf{0.3447} & \textbf{0.2043} & \textbf{1.1174} \\
\bottomrule
\end{tabularx}
\par}
\captionof{table}{Final grid distance (SG) on MM-GAG under \(5\times5\), \(B=10\), and \(C=4,\ldots,8\). A/G/T denote aerial-image, ground-view, and text targets. PPO, DiT, and AiRLoc are evaluated only for aerial-image targets.}
\label{tab:mmgag_sg}
\end{minipage}
\end{table*}

\paragraph{Experimental Setup.}
The main evaluation follows the protocol established by prior AGL benchmarks \cite{pirinen2023airloc,sarkar2024gomaa,mi2025geoexplorer}. The search region is discretized into a \(5\times5\) grid with initial Manhattan distances \(C=\{4,5,6,7,8\}\). We report success rate (SR) and final grid distance (SG), using greedy action selection with a legal-action mask. For \(N\) tasks, terminal cells \(p_i^T\), and goals \(p_i^g\),
\begin{equation}
\begin{aligned}
\mathrm{SR}&=\frac{1}{N}\sum_{i=1}^{N}\mathbf{1}[p_i^T=p_i^g],\\
\mathrm{SG}&=\frac{1}{N}\sum_{i=1}^{N}d_1(p_i^T,p_i^g),
\end{aligned}
\label{eq:eval_metrics}
\end{equation}
where \(\mathbf{1}[\cdot]\) is the indicator function and \(d_1\) is Manhattan distance. Thus SR is the fraction of searches that reach the goal within budget and SG is the mean terminal distance (lower is better). Tables~\ref{tab:mmgag_sr} and~\ref{tab:mmgag_sg} follow the MM-GAG setting; the additional studies use a unified \(B=12\), \(C=4,\ldots,8\) protocol with five trials per distance. The four-scenario \(B=12\) result is the unweighted macro-average over MASA~\cite{mnih2013machine}, MM-GAG, xBD-pre, and xBD-disaster~\cite{gupta2019xbd}.

Our final training pool combines 137 MASA images and 47 MM-GAG images. The model is trained for 480k PPO environment steps, dynamically sampling a start cell, goal cell, and initial distance \(C\in\{1,\ldots,8\}\) on each image rather than using fixed routes.

Evaluation uses the same source families as the AGL baselines: MASA for aerial-goal search, MM-GAG for aerial, ground-view, and text goals, GeoExplorer's SwissView100 and SwissViewMonuments splits~\cite{mi2025geoexplorer} for geographic-domain and unseen-target transfer, and xBD for disaster transfer. SwissView100 contains 100 Swiss aerial regions, while SwissViewMonuments contains 15 landmark or atypical-scene regions with aerial and ground-view cues. For xBD, we use an 800-pair test subset: xBD-pre uses pre-disaster targets and observations, whereas xBD-disaster keeps the pre-disaster target cue and searches post-disaster observations.

We compare with Random, PPO~\cite{schulman2017ppo} and DiT reimplementations, AiRLoc~\cite{pirinen2023airloc} where applicable, GOMAA-Geo~\cite{sarkar2024gomaa}, and GeoExplorer~\cite{mi2025geoexplorer}. Random samples legal actions uniformly; PPO uses the current observation without a history encoder; DiT follows offline optimal start--goal trajectories with a goal prefix and four actions \cite{sarkar2024gomaa,mi2025geoexplorer}. GOMAA-Geo supports all modalities, and GeoExplorer is the closest curiosity-driven baseline. In compact tables, GOMAA denotes GOMAA-Geo. Retrieval approaches \cite{arandjelovic2016netvlad,hu2018cvmnet,berton2022benchmark,zhu2021vigor,zhu2022transgeo} are omitted because they retrieve from a reference collection rather than act under the sequential protocol.

\begin{table}[!t]
\centering
{\small
\textbf{(a) Cross-scene, landmark, and disaster transfer}
\begin{tabularx}{\columnwidth}{@{}>{\raggedright\arraybackslash}X@{}Y@{}Y@{}Y@{}}
\toprule
Task & \multicolumn{3}{c}{SR\(\uparrow\) / SG\(\downarrow\)} \\
\cmidrule(lr){2-4}
 & GOMAA & GeoExplorer & \textbf{DynCur-Geo} \\
\midrule
MASA/A & 0.5640/1.3600 & 0.5600/1.2560 & \textbf{0.5880/1.1440} \\
xBD-pre/A & 0.5427/1.4132 & 0.5080/1.5555 & \textbf{0.5852/1.2922} \\
xBD-dis./A & 0.5345/1.4194 & 0.5129/1.5452 & \textbf{0.5856/1.2839} \\
Swiss100/A & 0.5044/1.5692 & 0.4700/1.5568 & \textbf{0.5772/1.3444} \\
SwissMon./A & 0.4596/1.6878 & 0.4604/1.7598 & \textbf{0.5208/1.5139} \\
SwissMon./G & 0.4464/1.8333 & 0.4593/1.7311 & \textbf{0.5084/1.5098} \\
\bottomrule
\end{tabularx}
\par
\textbf{(b) \(10\times10\) long-range extension}
\begin{tabularx}{\columnwidth}{@{}l@{\hspace{0.25em}}*{5}{Y@{}}}
\toprule
Method & \multicolumn{5}{c}{SR by initial distance \(C\)} \\
\cmidrule(lr){2-6}
 & 14 & 15 & 16 & 17 & 18 \\
\midrule
GOMAA & 0.5350 & 0.4750 & 0.6250 & 0.7500 & 0.7600 \\
GeoExplorer & 0.3350 & 0.3850 & 0.3350 & 0.2200 & 0.2250 \\
\textbf{DynCur-Geo} & \textbf{0.5400} & \textbf{0.6750} & \textbf{0.7550} & \textbf{0.8850} & \textbf{0.8850} \\
\bottomrule
\end{tabularx}
\par
\textbf{(c) Enlarged-grid summary}
\begin{tabularx}{\columnwidth}{@{}>{\raggedright\arraybackslash}p{0.31\columnwidth}@{}Y@{}Y@{}Y@{}}
\toprule
Setting & Method & SR\(\uparrow\) & SG\(\downarrow\) \\
\midrule
\multirow[c]{3}{*}{\shortstack[l]{\(8\times8,\ B=24\)\\\(C=10\text{--}14\)}} & GOMAA & 0.6790 & 1.3450 \\
 & GeoExplorer & 0.3680 & 3.0460 \\
 & \textbf{DynCur-Geo} & \textbf{0.7460} & \textbf{0.8590} \\
\midrule
\multirow[c]{3}{*}{\shortstack[l]{\(10\times10,\ B=32\)\\\(C=14\text{--}18\)}} & GOMAA & 0.6290 & 1.8010 \\
 & GeoExplorer & 0.3000 & 4.6770 \\
 & \textbf{DynCur-Geo} & \textbf{0.7480} & \textbf{0.7360} \\
\midrule
\multirow[c]{3}{*}{\shortstack[l]{\(25\times25,\ B=60\)\\\(C=12\text{--}48\)}} & GOMAA & 0.1800 & 13.5080 \\
 & GeoExplorer & 0.0820 & 15.5840 \\
 & \textbf{DynCur-Geo} & \textbf{0.2040} & \textbf{12.3120} \\
\bottomrule
\end{tabularx}
\par}
\caption{Transfer and long-range results. Panel (a) reports SR/SG under \(5\times5\), \(B=12\); GOMAA denotes GOMAA-Geo. Panel (b) reports SR and panel (c) reports SR/SG on the MASA aerial test split. The \(25\times25\) distances in panel (c) use a step size of four.}
\label{tab:transfer_longrange}
\end{table}

\begin{figure*}[!t]
\centering
\includegraphics[width=0.95\textwidth]{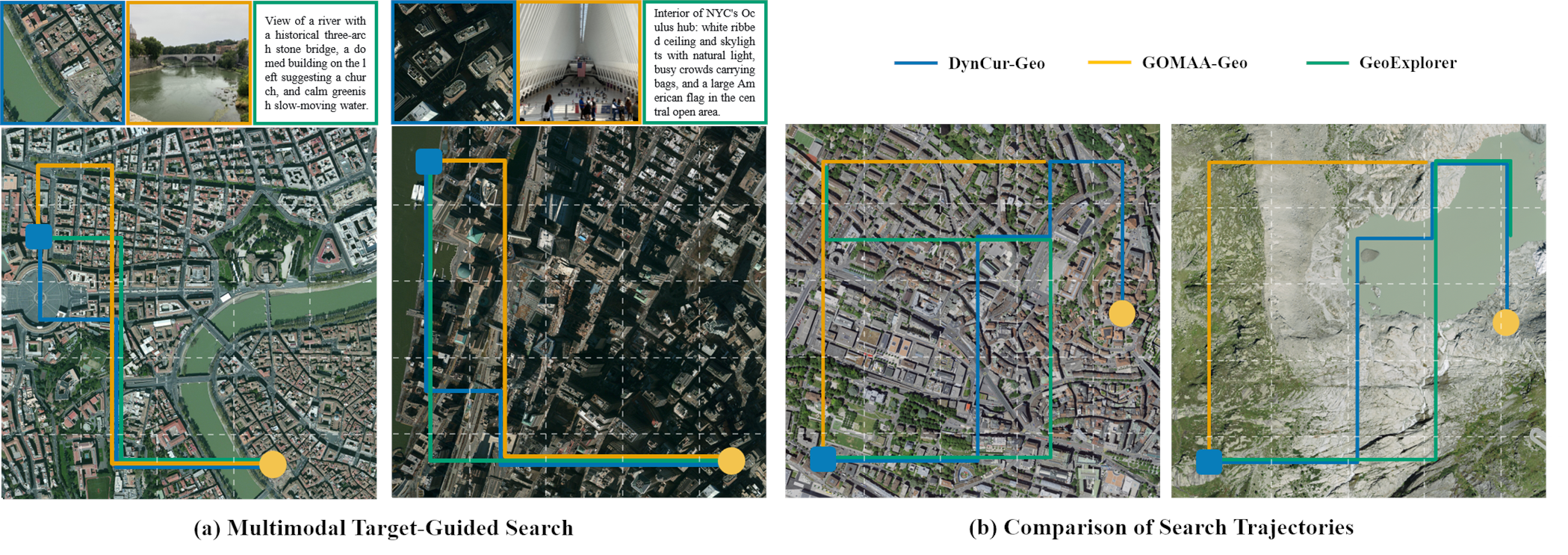}
\caption{Qualitative visualization of multimodal active geo-localization and test-time policy trajectories. In (a), blue, orange, and green target borders and their corresponding paths denote aerial-image, ground-view-image, and text cues, respectively. In (b), path colors identify DynCur-Geo (blue), GOMAA-Geo (orange), and GeoExplorer (green). Blue squares and yellow circles mark the start and goal cells.}
\label{fig:route_visualization}
\end{figure*}

\paragraph{MM-GAG Multimodal Results.}
Tables~\ref{tab:mmgag_sr} and~\ref{tab:mmgag_sg} report MM-GAG results for aerial-image, ground-view-image, and text cues; PPO, DiT, and AiRLoc apply only to aerial-target comparison. Our method achieves the highest average SR and lowest average SG in all three modalities, raising average SR over the strongest baseline from 0.5336 to 0.6170 for aerial targets, from 0.5523 to 0.6391 for ground-view targets, and from 0.5472 to 0.6247 for text targets.

The advantage is most pronounced at medium and long initial distances: several baselines remain stronger at \(C=4\), whereas our method leads clearly from \(C=6\) to \(C=8\). This pattern shows that the distance gate and PBRS are particularly effective when search requires broad exploration followed by accurate convergence; the lower SG further indicates closer endpoints on unsuccessful long-horizon episodes.

Because start--goal pairs are dynamically resampled during training, MM-GAG also tests unseen routes across all cue types; MASA, SwissView, and xBD further cover held-out scenes and appearance shifts.

\paragraph{Parameter Sensitivity.}
We vary the distance-gate lower bound \(\lambda_{\min}\) and PBRS coefficient \(\beta\) on nine evaluation settings. Table~\ref{tab:param_sensitivity} shows that \(\lambda_{\min}=0.405\) and \(\beta=0.10\) achieve the best joint SR/SG result. Both sweeps peak at interior settings: weaker or stronger gating, and removing or increasing PBRS, worsen at least one metric. This supports balanced distance-gated exploration and progress shaping.

\begin{table}[!t]
\centering
{\small
\textbf{(a) Distance-gate lower bound \(\lambda_{\min}\)}
\begin{tabularx}{\columnwidth}{@{}l@{\hspace{0.25em}}*{5}{Y@{}}}
\toprule
\(\lambda_{\min}\) & 0.100 & 0.250 & 0.405 & 0.650 & 0.900 \\
\midrule
Avg. SR\(\uparrow\) & 0.4813 & 0.5009 & \textbf{0.5765} & 0.5089 & 0.4582 \\
Avg. SG\(\downarrow\) & 1.9150 & 1.7052 & \textbf{1.2885} & 1.6526 & 1.9135 \\
\bottomrule
\end{tabularx}
\par
\textbf{(b) PBRS coefficient \(\beta\)}
\begin{tabularx}{\columnwidth}{@{}l@{\hspace{0.25em}}*{5}{Y@{}}}
\toprule
\(\beta\) & 0.00 & 0.05 & 0.10 & 0.15 & 0.20 \\
\midrule
Avg. SR\(\uparrow\) & 0.5535 & 0.5134 & \textbf{0.5765} & 0.4933 & 0.5307 \\
Avg. SG\(\downarrow\) & 1.3290 & 1.5409 & \textbf{1.2885} & 1.7120 & 1.4880 \\
\bottomrule
\end{tabularx}
\par}
\caption{Sensitivity to reward-shaping parameters under \(5\times5\), \(B=10\), and \(C=4,\ldots,8\), macro-averaged over nine MASA, MM-GAG, SwissView, and xBD evaluation settings.}
\label{tab:param_sensitivity}
\end{table}

\paragraph{Qualitative Path Visualization.}
Fig.~\ref{fig:route_visualization} illustrates both multimodal cue generality and trajectory quality. Our method makes broader early progress and becomes goal-directed near the target, whereas competing policies often revisit nearby cells or drift after plausible local moves. The comparison is consistent with the quantitative distance-wise pattern: the benefit of the proposed policy is most visible when early exploratory moves must be followed by a decisive final approach.

\begin{figure}[!t]
\centering
\includegraphics[width=0.82\columnwidth]{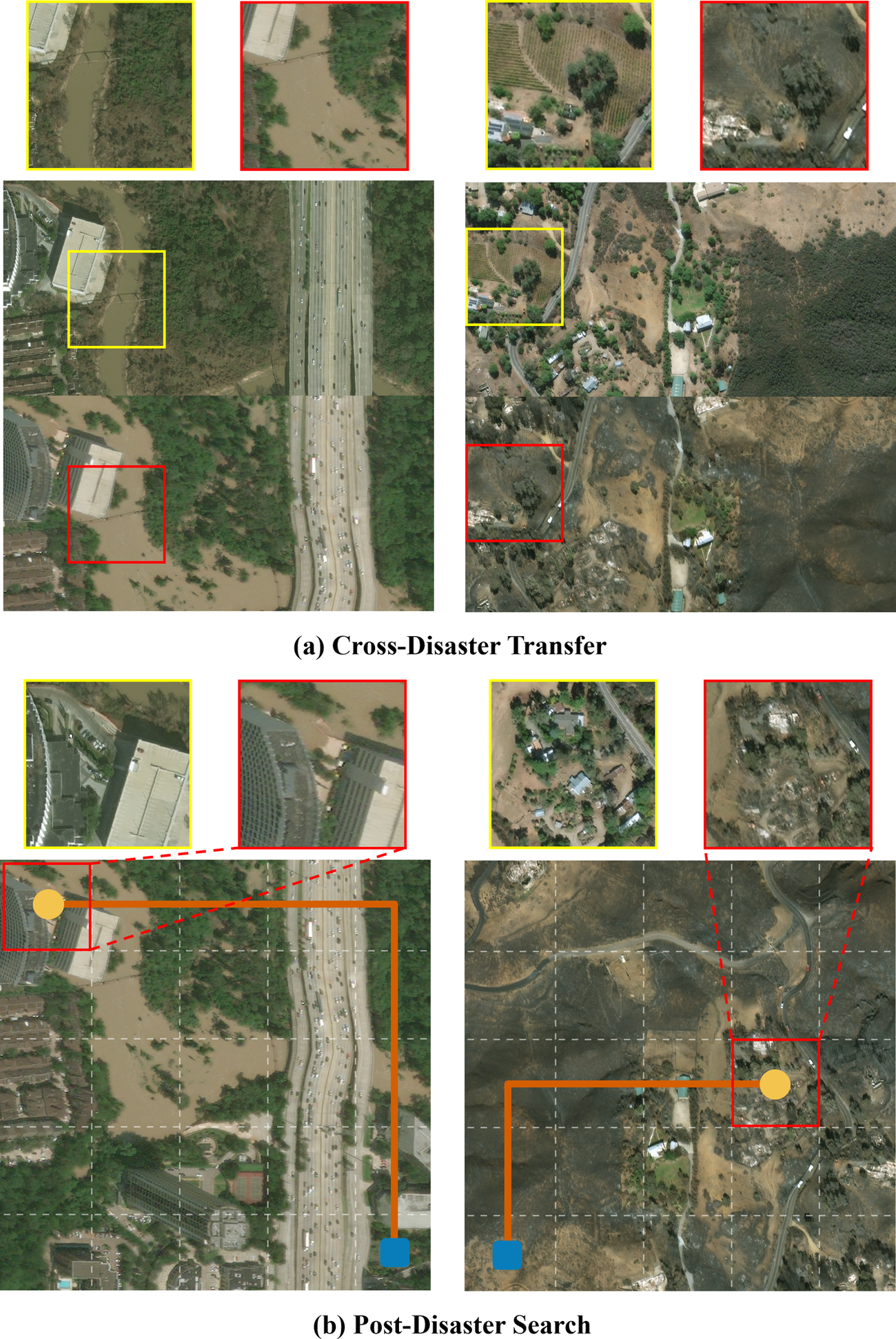}
\caption{Qualitative xBD cross-disaster visualization. Yellow boxes indicate pre-disaster target regions and red boxes indicate the corresponding post-disaster regions. The lower panels show the pre-disaster target cue and the post-disaster search scene with a test-time policy trajectory; blue squares and yellow circles mark the start and goal cells.}
\label{fig:xbd_visualization}
\end{figure}

\paragraph{Cross-Scene and Disaster Transfer.}
Table~\ref{tab:transfer_longrange}(a) shows the best SR/SG pair for our method across MASA, SwissView, and xBD. The xBD-disaster setting uses a pre-disaster target cue with post-disaster search observations (Fig.~\ref{fig:xbd_visualization}), while SwissViewMonuments tests landmark-centered held-out scenes. Our method improves both metrics on the two xBD splits, rather than trading more successes for worse failed trajectories, and the SwissView results extend the gain to held-out landmark cues.

\paragraph{Long-Range Extension.}
Table~\ref{tab:transfer_longrange}(b,c) further evaluates enlarged MASA aerial grids. On \(10\times10\) with \(B=32\) and \(C=\{14,15,16,17,18\}\), our method obtains 0.7480 SR and 0.7360 SG, compared with 0.6290 and 1.8010 for GOMAA-Geo, and leads every distance bin. Our method also retains the highest SR on \(8\times8\) and \(25\times25\), although all methods decline in the most demanding setting. Its consistently lower SG shows that failed episodes end closer to the target across grid scales.

\paragraph{Unified \(B=12\) Main Comparison.}
Table~\ref{tab:main_b12_extra} compares MASA, MM-GAG, xBD-pre, and xBD-disaster under \(B=12\). Our method achieves the highest mean SR (0.6129) and leads all four settings. This comparison separates the gain from any single benchmark: the improvement appears on the standard aerial setting, multimodal aerial-target evaluation, and both disaster-related conditions under one common budget. The consistent ranking across aerial, multimodal, and disaster-transfer settings suggests that the reward design improves the search policy itself, rather than only exploiting a particular target modality or scene split. Notably, the gains on xBD-pre and xBD-disaster are close, indicating that the learned policy does not depend on unchanged visual appearance between the target cue and search scene.

\begin{table}[!t]
\centering
{\small
\setlength{\tabcolsep}{2.0pt}
\begin{tabularx}{\columnwidth}{@{}l@{\hspace{0.25em}}*{5}{Y@{}}}
\toprule
Method & MASA & MM-GAG & xBD-pre & xBD-dis. & Avg. \\
\midrule
Random & 0.1000 & 0.0868 & 0.0925 & 0.0925 & 0.0929 \\
PPO & 0.2280 & 0.1991 & 0.1712 & 0.1676 & 0.1915 \\
DiT & 0.2280 & 0.2255 & 0.2409 & 0.2365 & 0.2327 \\
GOMAA-Geo & 0.5800 & 0.5464 & 0.5612 & 0.5534 & 0.5602 \\
GeoExplorer & 0.5640 & 0.5421 & 0.5233 & 0.5372 & 0.5417 \\
\textbf{DynCur-Geo} & \textbf{0.6200} & \textbf{0.6357} & \textbf{0.5973} & \textbf{0.5984} & \textbf{0.6129} \\
\bottomrule
\end{tabularx}
\par}
\caption{Unified \(B=12\) SR on MASA/A, MM-GAG/A, xBD-pre/A, and xBD-disaster/A; Avg. denotes their macro-average.}
\label{tab:main_b12_extra}
\end{table}

\paragraph{Budget Sensitivity.}
Table~\ref{tab:budget_sensitivity} sweeps the search budget from \(B=4\) to \(B=12\). With \(C=4,\ldots,8\), the smallest budgets leave some distance groups geometrically unreachable.

\begin{table}[!t]
\centering
{\small
\setlength{\tabcolsep}{2.0pt}
\textbf{(a) Success rate (SR)}
\begin{tabularx}{\columnwidth}{@{}l@{\hspace{0.25em}}*{5}{Y@{}}}
\toprule
Method & \(B=4\) & \(B=6\) & \(B=8\) & \(B=10\) & \(B=12\) \\
\midrule
GOMAA & 0.0440 & 0.1791 & 0.4891 & 0.5424 & 0.5602 \\
GeoExplorer & \textbf{0.0459} & \textbf{0.1842} & 0.4805 & 0.5281 & 0.5417 \\
\textbf{DynCur-Geo} & 0.0366 & 0.1790 & \textbf{0.5326} & \textbf{0.5917} & \textbf{0.6129} \\
\bottomrule
\end{tabularx}
\par
\textbf{(b) Final grid distance (SG)}
\begin{tabularx}{\columnwidth}{@{}l*{5}{Y}@{}}
\toprule
Method & \(B=4\) & \(B=6\) & \(B=8\) & \(B=10\) & \(B=12\) \\
\midrule
GOMAA & 3.2143 & 1.9513 & 1.4263 & 1.4138 & 1.4324 \\
GeoExplorer & 3.2261 & 1.9501 & 1.4453 & 1.4550 & 1.4984 \\
\textbf{DynCur-Geo} & \textbf{2.8507} & \textbf{1.8584} & \textbf{1.2760} & \textbf{1.2304} & \textbf{1.2128} \\
\bottomrule
\end{tabularx}
\par}
\captionof{table}{Budget sensitivity (SR/SG), macro-averaged over MASA/A, MM-GAG/A, xBD-pre/A, and xBD-disaster/A.}
\label{tab:budget_sensitivity}
\end{table}

\begin{table}[!t]
\centering
{\small
\begin{tabularx}{\columnwidth}{@{}>{\centering\arraybackslash}p{0.07\columnwidth}@{}>{\centering\arraybackslash}p{0.07\columnwidth}@{}G@{}>{\centering\arraybackslash}p{0.10\columnwidth}@{}*{5}{Y@{}}}
\toprule
Ext. & Int. & Gate & PBRS & \(C=4\) & \(C=5\) & \(C=6\) & \(C=7\) & \(C=8\) \\
\midrule
\(\surd\) &  &  &  & 0.3617 & \textbf{0.4326} & 0.5801 & 0.7305 & 0.7773 \\
\(\surd\) &  &  & \(\surd\) & 0.3248 & 0.3546 & 0.6085 & 0.8014 & 0.8823 \\
 & \(\surd\) & 1 &  & 0.1518 & 0.1064 & 0.0794 & 0.0752 & 0.0468 \\
\(\surd\) & \(\surd\) & 1 &  & 0.2525 & 0.3418 & 0.5816 & 0.8184 & 0.8369 \\
\(\surd\) & \(\surd\) & Const &  & \textbf{0.3787} & 0.3972 & 0.5220 & 0.5929 & 0.5787 \\
\(\surd\) & \(\surd\) & Const & \(\surd\) & 0.3050 & 0.4099 & 0.5986 & 0.7816 & 0.8426 \\
\(\surd\) & \(\surd\) & Sin &  & 0.3489 & 0.4241 & 0.4780 & 0.4028 & 0.2780 \\
\(\surd\) & \(\surd\) & Sin & \(\surd\) & 0.3177 & 0.3702 & 0.4879 & 0.4766 & 0.4596 \\
\(\surd\) & \(\surd\) & Power &  & 0.2879 & 0.3475 & 0.4496 & 0.5986 & 0.6468 \\
\(\surd\) & \(\surd\) & Power & \(\surd\) & 0.2525 & 0.3291 & 0.5191 & 0.6809 & 0.7660 \\
\(\surd\) & \(\surd\) & Line &  & 0.2695 & 0.3716 & 0.5787 & 0.8482 & 0.8426 \\
\(\surd\) & \(\surd\) & Line & \(\surd\) & 0.2709 & 0.4156 & \textbf{0.6638} & \textbf{0.8681} & \textbf{0.9163} \\
\bottomrule
\end{tabularx}
\par}
\captionof{table}{Reward ablation on MM-GAG/A/G/T (\(5\times5\), \(B=10\)); each \(C\) column averages A/G/T.}
\label{tab:reward_gate_ablation}
\end{table}

From \(B=8\) onward, our method consistently attains the highest SR and lowest SG. In particular, SR rises from 0.5326 at \(B=8\) to 0.6129 at \(B=12\), while SG decreases from 1.2760 to 1.2128, which confirms that more available actions improve both successful localization and the quality of failed episodes.
At \(B=12\), this corresponds to a 0.0527 SR gain and a 0.2196 SG reduction relative to the strongest competing entry for each metric.
This budget trend is important because smaller budgets mainly test whether the policy avoids wasteful wandering, whereas larger budgets test whether early exploration can be converted into final convergence.
The simultaneous SR and SG improvements therefore indicate that dynamic weighting does not merely produce more isolated successes; it also leaves unsuccessful trajectories closer to the target.

\paragraph{Reward and Gate Ablation.}
Table~\ref{tab:reward_gate_ablation} evaluates reward combinations on MM-GAG over aerial-image, ground-view, and text cues. Each \(C\) column averages SR over the three target forms under \(5\times5\), \(B=10\). External reward is essential, while the complete linear-gate and PBRS formulation is strongest at \(C=6\) to \(C=8\), improving from 0.2709 at \(C=4\) to 0.9163 at \(C=8\); intrinsic reward alone falls to 0.0468 at \(C=8\). PBRS raises the mean SR for every gated schedule, but the linear gate remains best after shaping and leads at \(C=6\) to \(C=8\). This pattern indicates that PBRS supplies broadly useful progress feedback, whereas the gate determines how effectively curiosity is allocated over the search horizon. In other words, curiosity is beneficial when it is treated as a progress-aware training signal, not as a static exploration bonus.

\section{Conclusion}

This paper studied how curiosity should be allocated in multimodal active geo-localization, where sparse feedback, limited local observations, and heterogeneous target cues make search inefficient. We introduced distance-gated curiosity reward shaping, which modulates prediction-error intrinsic reward by remaining target distance and combines it with potential-based progress shaping. This design preserves exploratory behavior when the target is far away while reducing novelty-seeking pressure near the goal. Across multimodal, cross-scene, disaster-affected, and enlarged-grid settings, the resulting policy improves both success rate and final grid distance over the evaluated active geo-localization baselines. These results suggest that curiosity is most useful for AGL when conditioned on task progress, and they motivate future extensions to continuous UAV dynamics, richer map scales, and real deployment constraints.
\bibliography{aaai2027_refs}
\clearpage
\setcounter{section}{0}
\setcounter{subsection}{0}
\setcounter{figure}{0}
\setcounter{table}{0}
\setcounter{equation}{0}
\renewcommand{\thesection}{S\arabic{section}}
\renewcommand{\thesubsection}{S\arabic{section}.\arabic{subsection}}
\renewcommand{\thefigure}{S\arabic{figure}}
\renewcommand{\thetable}{S\arabic{table}}
\renewcommand{\theequation}{S\arabic{equation}}
\setcounter{topnumber}{4}
\setcounter{bottomnumber}{2}
\setcounter{totalnumber}{8}
\setcounter{dbltopnumber}{4}
\renewcommand{\topfraction}{0.98}
\renewcommand{\bottomfraction}{0.90}
\renewcommand{\textfraction}{0.02}
\renewcommand{\floatpagefraction}{0.88}
\renewcommand{\dbltopfraction}{0.98}
\renewcommand{\dblfloatpagefraction}{0.88}
\setlength{\textfloatsep}{6pt plus 1pt minus 2pt}
\setlength{\dbltextfloatsep}{6pt plus 1pt minus 2pt}
\setlength{\floatsep}{5pt plus 1pt minus 2pt}
\setlength{\dblfloatsep}{5pt plus 1pt minus 2pt}
\setlength{\abovecaptionskip}{3pt}
\setlength{\belowcaptionskip}{0pt}
\section*{Supplementary Materials}
\section{Scope and Diagnostic Protocol}

The supplementary material reports additional trajectory-level results for the fixed evaluation protocols used in the main paper. It includes route visualizations, route-dynamics summaries, cross-target evaluations, enlarged-grid stress tests, and implementation details for evaluation-time rollout. Aggregate values are reported when they identify the scope of a route set or connect a qualitative pattern to the corresponding evaluator. The analyses use the same data families as the main paper: MASA, MM-GAG, GeoExplorer's SwissView100 and SwissViewMonuments splits, and xBD.

\begin{figure*}[t]
\centering
\includegraphics[width=\textwidth]{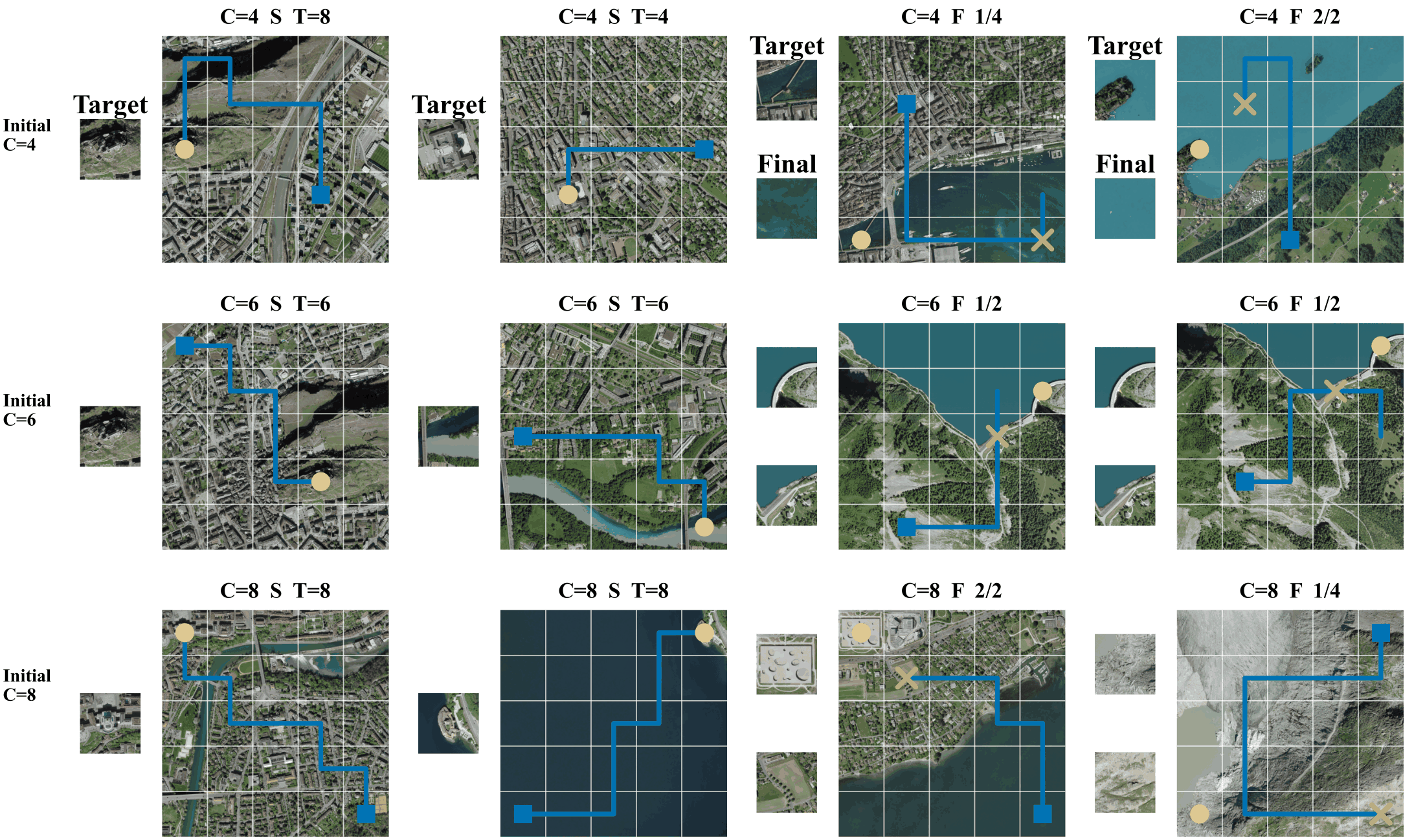}
\caption{Distance-stratified route gallery for \methodname{}. Each row contains two successes followed by two failures at the same initial distance. Every side cue rail shows the target photograph; failed panels also show the final-cell photograph. $S/F$ denotes success/failure, $T$ is path length for a success, and the pair on a failure is minimum/final distance.}
\label{fig:distance_gallery}
\end{figure*}

\begin{figure*}[t]
\centering
\includegraphics[width=\textwidth]{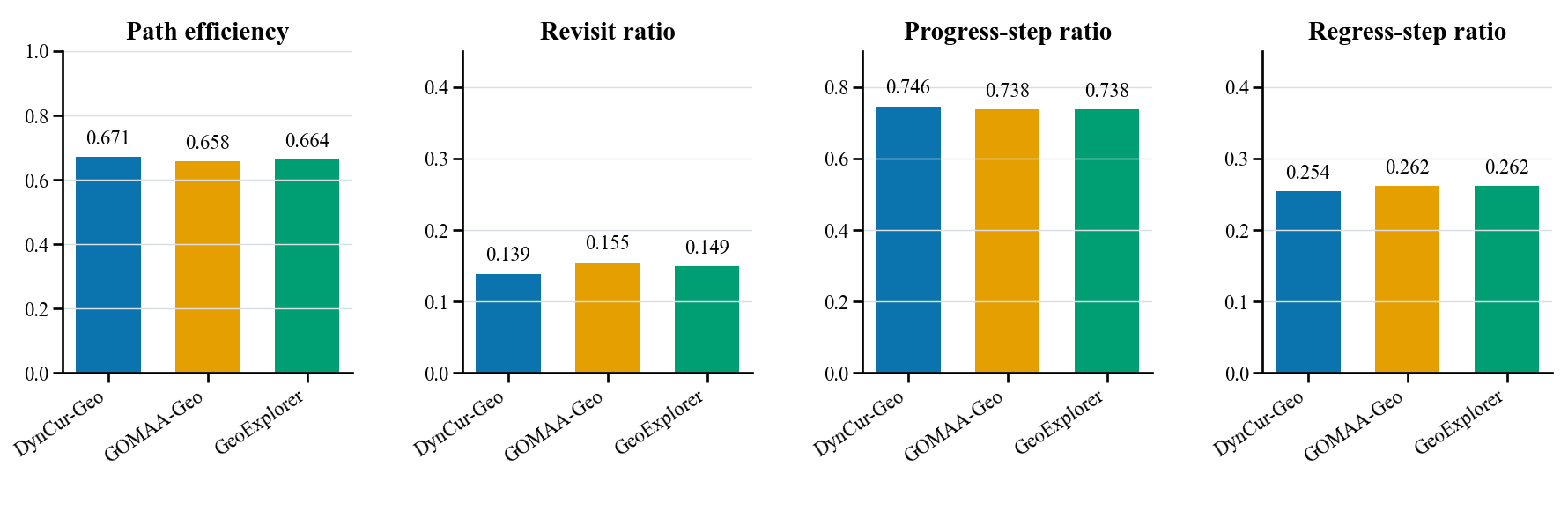}
\caption{Aggregate route-shape descriptors on all 735 shared diagnostic tasks per policy. DynCur-Geo denotes the proposed method, matching the main-paper terminology. The method has slightly higher path efficiency and progress-step ratio, and slightly lower revisit and regress ratios, than the two baselines.}
\label{fig:behavior_profile}
\end{figure*}

The material is organized to separate aggregate evidence from route-level interpretation. Table~\ref{tab:diagnostic_distance} and Tables~\ref{tab:training_pool_selection}--\ref{tab:long_range_25} provide the main quantitative additions: shared-route descriptors, training-pool selection, distance-stratified results, development hyperparameter checks, and enlarged-grid stress tests. Figures~\ref{fig:distance_gallery}--\ref{fig:population_dynamics} then explain how these scores arise from successful routes, persistent failure modes, matched same-task contrasts, realized reward traces, and step-level reward distributions. Figures~\ref{fig:target_modalities}--\ref{fig:long_range} extend the same analysis to multimodal targets, xBD pre/post search, Swiss ground-view targets, and larger search regions.

\subsection{Recorded Route Bank}

The principal trajectory bank contains 2,205 SwissViewMonuments aerial-goal trajectories. It consists of 735 shared task keys evaluated by \methodname{}, GOMAA-Geo, and GeoExplorer. A key fixes the aerial scene, start cell, target cell, initial Manhattan distance, and repeat index, so all three trajectories within a key solve the same navigation problem with the same target cue. The bank uses a $5\times5$ grid, action budget $B=10$, initial distances $C\in\{4,6,8\}$, a fixed task-bank seed, greedy action selection, and legal-action masking. It contains 375 tasks at $C=4$, 300 at $C=6$, and 60 at $C=8$ for each policy. Successful episodes terminate on reaching the target; unsuccessful episodes exhaust the ten-action budget. This supplement-only diagnostic bank is distinct from the $B=12$ SwissViewMonuments transfer evaluation reported in the main paper.

The route bank is used for controlled path inspection. When methods are compared, the image, start cell, target cell, initial distance, repeat index, and action budget are identical. Unless another evaluator is specified, success rates, paired counts, and case frequencies in this supplement refer to these 735 shared tasks. Route galleries are grouped by behavior and matched outcome.

\subsection{Route Descriptors}

For a route $\tau=(p_0,\ldots,p_T)$, path efficiency is $C/T$, capped naturally at one because $C$ is the shortest grid distance. A revisited step enters a cell seen earlier in the same route. An immediate backtrack satisfies $p_t=p_{t-2}$. Progress and regress steps respectively reduce and increase $d_1(p_t,p_g)$. The target neighborhood is the set of cells with distance at most one. Post-entry drift counts distance-increasing actions after first entering this neighborhood.

Successful routes are grouped as efficient success ($T=C$), detour success, or late convergence. Failed routes are grouped as post-entry drift, revisit/ping-pong, approach-then-regress, non-convergent, or partial progress. These deterministic descriptors support consistent visualization across figures. Blue squares denote starts, yellow circles targets, and yellow crosses failed terminal cells. Route colors are fixed throughout: blue for \methodname{}, orange for GOMAA-Geo, and green for GeoExplorer.

Each single-policy route panel places the target photograph in a compact side cue rail; failed panels also include the final-cell photograph. Each matched three-policy task uses one shared target photograph beside the route set. The photographs are grid-cell crops aligned with the aerial scene, while the full $5\times5$ route image remains uncropped and unobstructed. Failures are interpreted relative to the requested target cell rather than as generic spatial coverage.

\begin{table}[!t]
\centering
{\suppnarrowtable
\renewcommand{\arraystretch}{1.0}
\textbf{(a) Outcome and path-efficiency descriptors}
\begin{tabularx}{\columnwidth}{@{}>{\raggedright\arraybackslash}p{0.25\columnwidth}>{\centering\arraybackslash}p{0.07\columnwidth}*{4}{Y}@{}}
\toprule
Method & $C$ & Tasks & SR & SG & Eff. \\
\midrule
\multirow{3}{*}{\methodname} & 4 & 375 & 0.2507 & 2.5867 & 0.5279 \\
 & 6 & 300 & 0.6067 & 1.0867 & 0.7927 \\
 & 8 & 60 & 0.8500 & 0.4000 & 0.9600 \\
\midrule
\multirow{3}{*}{GOMAA-Geo} & 4 & 375 & 0.2507 & 2.5653 & 0.5179 \\
 & 6 & 300 & 0.5900 & 1.0800 & 0.7782 \\
 & 8 & 60 & 0.6833 & 0.8667 & 0.9333 \\
\midrule
\multirow{3}{*}{GeoExplorer} & 4 & 375 & 0.3333 & 2.2453 & 0.5464 \\
 & 6 & 300 & 0.4733 & 1.5400 & 0.7587 \\
 & 8 & 60 & 0.6000 & 1.1333 & 0.9200 \\
\bottomrule
\end{tabularx}
\par
\vspace{1pt}
\textbf{(b) Route-dynamics descriptors}
\begin{tabularx}{\columnwidth}{@{}>{\raggedright\arraybackslash}p{0.25\columnwidth}>{\centering\arraybackslash}p{0.07\columnwidth}*{3}{Y}@{}}
\toprule
Method & $C$ & Revisit & Prog. & Reg. \\
\midrule
\multirow{3}{*}{\methodname} & 4 & 0.1903 & 0.6346 & 0.3654 \\
 & 6 & 0.0915 & 0.8420 & 0.1580 \\
 & 8 & 0.0500 & 0.9600 & 0.0400 \\
\midrule
\multirow{3}{*}{GOMAA-Geo} & 4 & 0.1912 & 0.6307 & 0.3693 \\
 & 6 & 0.1216 & 0.8351 & 0.1649 \\
 & 8 & 0.0967 & 0.9233 & 0.0767 \\
\midrule
\multirow{3}{*}{GeoExplorer} & 4 & 0.1648 & 0.6609 & 0.3391 \\
 & 6 & 0.1337 & 0.8023 & 0.1977 \\
 & 8 & 0.1283 & 0.9033 & 0.0967 \\
\bottomrule
\end{tabularx}
\par}
\caption{Distance-stratified descriptors on the 735 shared diagnostic tasks. SR is exact-hit success rate and SG is terminal grid distance. Eff. abbreviates path efficiency; revisit, progress, and regress are route-level means. These values characterize the route bank used for the following visual analysis.}
\label{tab:diagnostic_distance}
\end{table}

Table~\ref{tab:diagnostic_distance} summarizes the distance-conditioned route statistics used by the galleries. At $C=4$, the main differences occur near the target. At $C=6$, \ourmethod{} has similar SG to GOMAA-Geo with higher efficiency and fewer revisits. At $C=8$, \ourmethod{} reaches 0.8500 SR with 0.9600 path efficiency, compared with 0.6833 and 0.6000 SR for the two baselines. The three distance groups have different sample counts, so comparisons are made within each stratum.

\section{Successful Route Mechanisms}

\begin{figure*}[t]
\centering
\includegraphics[width=\textwidth]{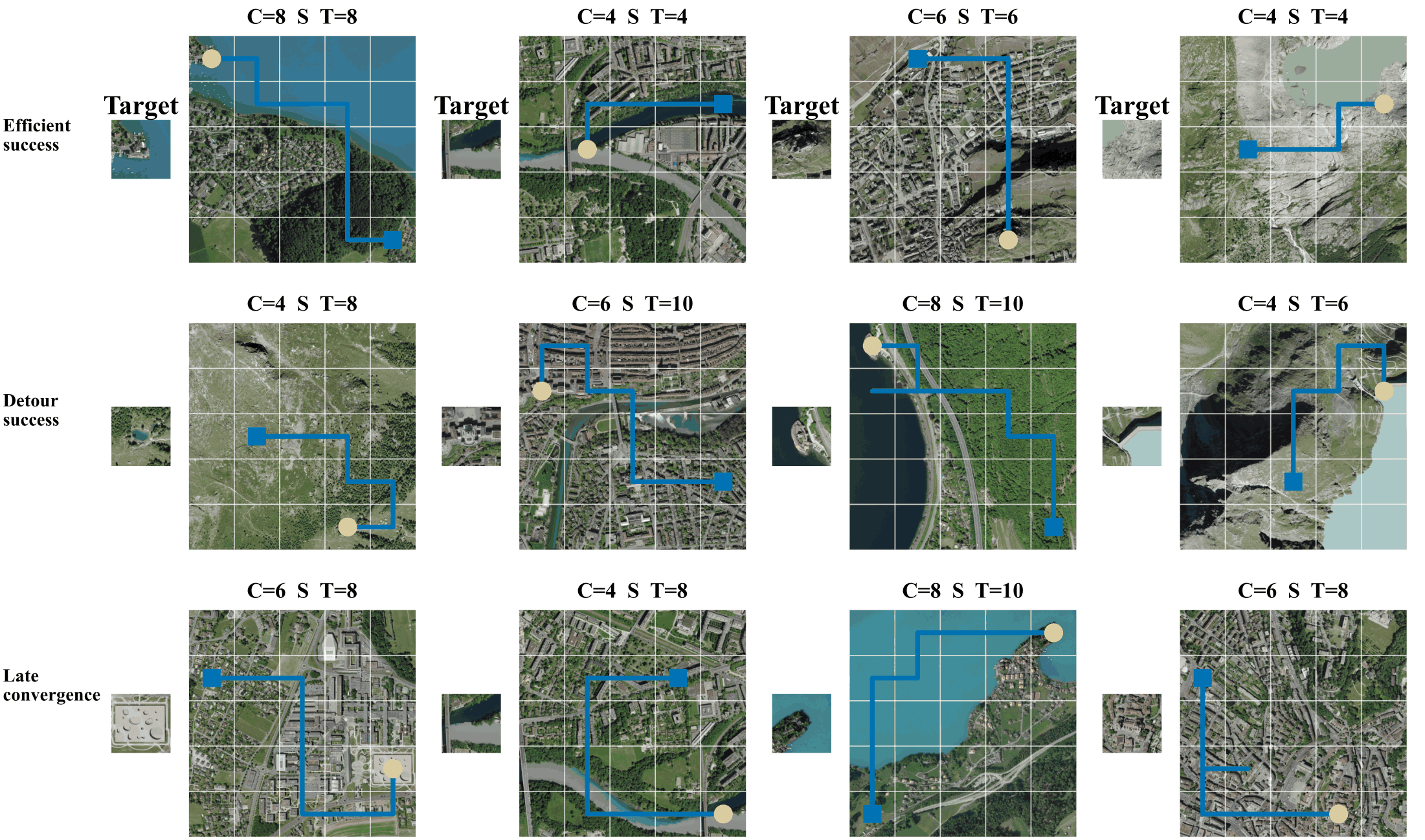}
\caption{Successful-route atlas for \methodname{}: four efficient successes, four detour successes, and four late-convergence successes. $C$ is initial distance, $S$ marks success, and $T$ is path length. Each panel pairs the route image with its target-cell photograph in a compact side cue rail.}
\label{fig:success_atlas}
\end{figure*}

Fig.~\ref{fig:success_atlas} contains 12 recorded successes from three behavior groups. The gallery separates three ways in which a successful policy can use its action budget. Efficient successes follow a shortest path and provide the cleanest visible expression of target-directed search. Detour successes temporarily preserve or increase target distance but recover within the action budget. Late-convergence successes spend most of the route outside the target neighborhood and enter it near termination.

Across all 327 successful routes of \ourmethod{} in the shared bank, mean path efficiency is 0.9265, the mean revisit ratio is 0.0086, and immediate backtracks average 0.0795 per route. The corresponding successful-route efficiencies are 0.9046 for GOMAA-Geo and 0.9033 for GeoExplorer. The qualitative panels show the same pattern: most routes of \ourmethod{} make monotonic progress, with only a small number of corrective moves in the detour and late-convergence rows.

The efficient row shows target-directed motion with little route correction. The detour row contains temporary regress actions followed by recovery and termination. The late-convergence row shows broader motion while the target remains distant, followed by contraction near the end of the route. These patterns are consistent with the reward design, but a successful path alone does not isolate individual reward terms; the stepwise reward diagnostics below report the corresponding traces.

Of the 254 efficient successes in the recorded bank, 75 occur at $C=4$, 131 at $C=6$, and 48 at $C=8$. Detour and late-convergence successes appear at all recorded distances. The resulting behavior is not a fixed explore-then-exploit schedule. Because the distance weight is state dependent, a regress action can raise the intrinsic weight again, and convergence can occur at different points in the route.

\section{Failure Taxonomy and Mechanism Boundaries}

\begin{figure*}[t]
\centering
\includegraphics[width=\textwidth]{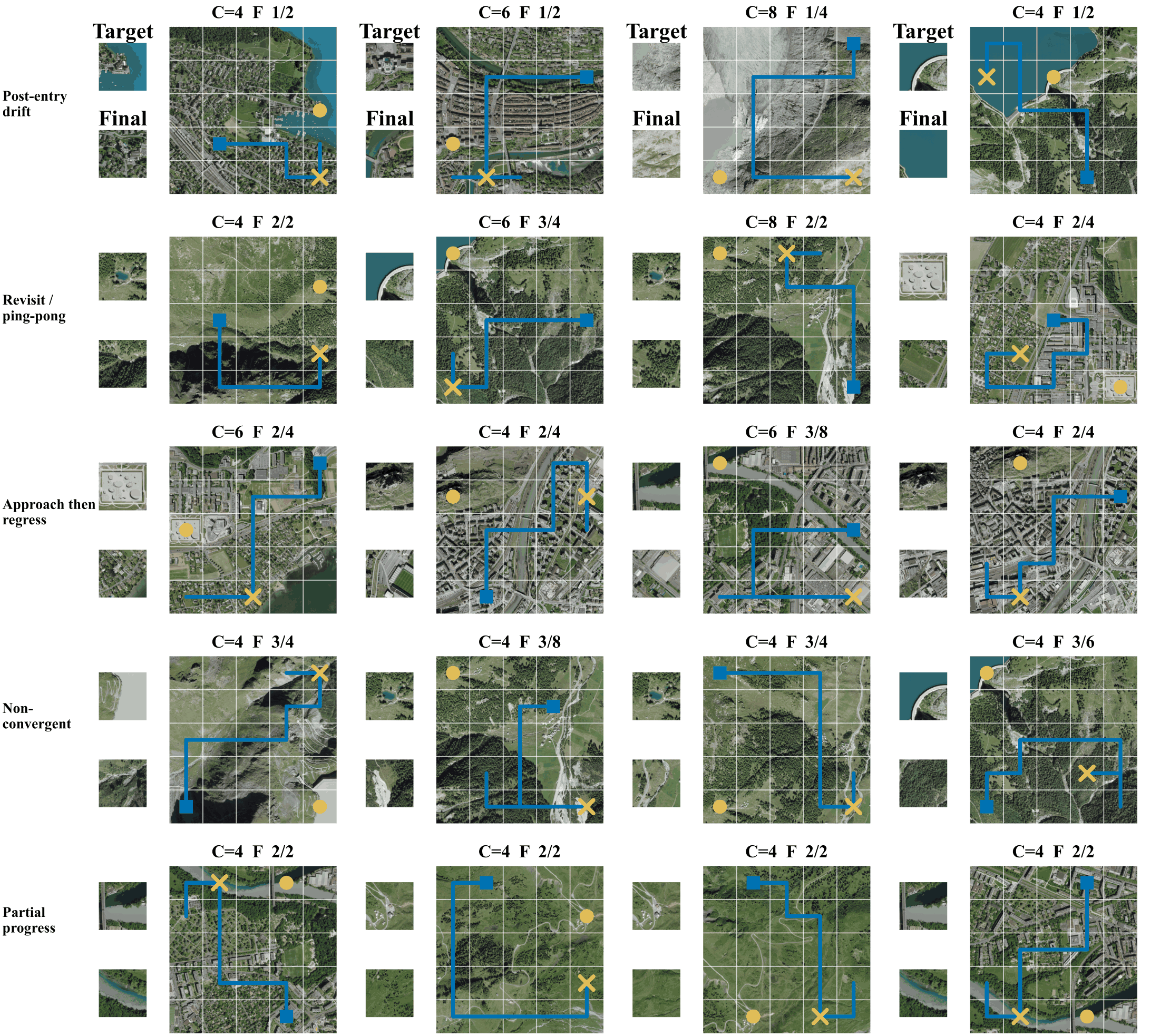}
\caption{Failure atlas for \methodname{}. Rows show four examples each of post-entry drift, revisit/ping-pong, approach-then-regress, non-convergence, and partial progress. Each compact side cue rail compares target and final-cell photographs, while a yellow cross marks the final cell in the full route view. $F$ marks failure; the numeric pair is minimum/final target distance.}
\label{fig:failure_atlas}
\end{figure*}

Fig.~\ref{fig:failure_atlas} contains 20 failures across five route categories. The examples include short-, medium-, and long-distance routes; urban, mountain, forest, shoreline, and water-dominated scenes; and both near-target and globally divergent endings. Target and terminal cues are shown together so that each failure can be interpreted as an endpoint error, a loop, or a global search error.

The full recorded bank contains 408 failures of \ourmethod{}. Post-entry drift accounts for 220 routes (53.9\% of failures), and revisit/ping-pong accounts for 133 (32.6\%). Together they cover 86.5\% of failures. The remaining failures comprise 36 approach-then-regress, 12 non-convergent, and 7 partial-progress routes. Many failed routes reduce distance substantially before losing the correct endpoint.

\paragraph{Post-entry drift.}
These routes reach distance one and subsequently move away. In the top row of Fig.~\ref{fig:failure_atlas}, the trajectory enters the goal neighborhood, but the policy selects an adjacent visually plausible cell or leaves the neighborhood after a local ambiguity. Distance gating reduces intrinsic weight near the target while preserving a small residual exploration weight. Final-cell confirmation is handled by the next policy action under the fixed action budget.

\paragraph{Revisit and ping-pong.}
These failures repeatedly enter previously visited cells or reverse the immediately preceding action. Prediction-error curiosity is model-relative rather than a direct count of spatial coverage, and repeated observations can still have nonzero prediction error. Recurrent history reduces, but does not eliminate, repeated-cell behavior. The distance weight changes the magnitude of the intrinsic term but does not prohibit return transitions.

\paragraph{Approach then regress.}
These routes reduce distance to at most half of the initial value and later give back at least two cells of progress. When a route moves away from the target, the remaining-distance ratio and intrinsic weight rise again. This behavior supports recovery from an incorrect local hypothesis, but it also makes endpoint retention depend on the learned visual state.

\paragraph{Non-convergence and partial progress.}
Non-convergent routes finish at least as far away as they started; partial-progress routes improve SG without entering the target neighborhood. All 19 such failures in the recorded bank occur at $C=4$. This concentration matches the short-range weakness in Table~\ref{tab:diagnostic_distance}: when the target is initially close, a single wrong endpoint decision consumes a large fraction of the budget.

The paired target cues expose several plausible ambiguities, including repeated road blocks, homogeneous forest, water boundaries, and snow-covered terrain. These observations are used as qualitative context for route-level errors; quantitative claims are made from route descriptors and aggregate evaluator outputs.

\section{Matched Same-Task Contrasts}

The paired bank permits controlled qualitative comparison with scene, target cue, start cell, target cell, distance, repeat, and action budget fixed. Table~\ref{tab:outcome_patterns} reports every success combination across the 735 shared keys. The proposed policy succeeds alone on 89 tasks, while at least one baseline succeeds on 224 tasks where it fails. The two directions are both visualized.

\begin{table}[t]
\centering
{\suppnarrowtable
\begin{tabularx}{\columnwidth}{@{}>{\raggedright\arraybackslash}p{0.41\columnwidth}@{}*{4}{Y}@{}}
\toprule
Successful methods & $C=4$ & $C=6$ & $C=8$ & Total \\
\midrule
None & 155 & 29 & 0 & 184 \\
GeoExplorer only & 57 & 15 & 1 & 73 \\
GOMAA-Geo only & 44 & 41 & 2 & 87 \\
GOMAA-Geo + GeoExplorer & 25 & 33 & 6 & 64 \\
DynCur-Geo only & 40 & 39 & 10 & 89 \\
DynCur-Geo + GeoExplorer & 29 & 40 & 8 & 77 \\
DynCur-Geo + GOMAA-Geo & 11 & 49 & 12 & 72 \\
All three & 14 & 54 & 21 & 89 \\
\bottomrule
\end{tabularx}
\par}
\caption{Success-combination counts on shared diagnostic tasks. Each row lists the methods that hit the target within $B=10$; omitted methods failed on the same task.}
\label{tab:outcome_patterns}
\end{table}

\begin{figure*}[t]
\centering
\includegraphics[width=\textwidth]{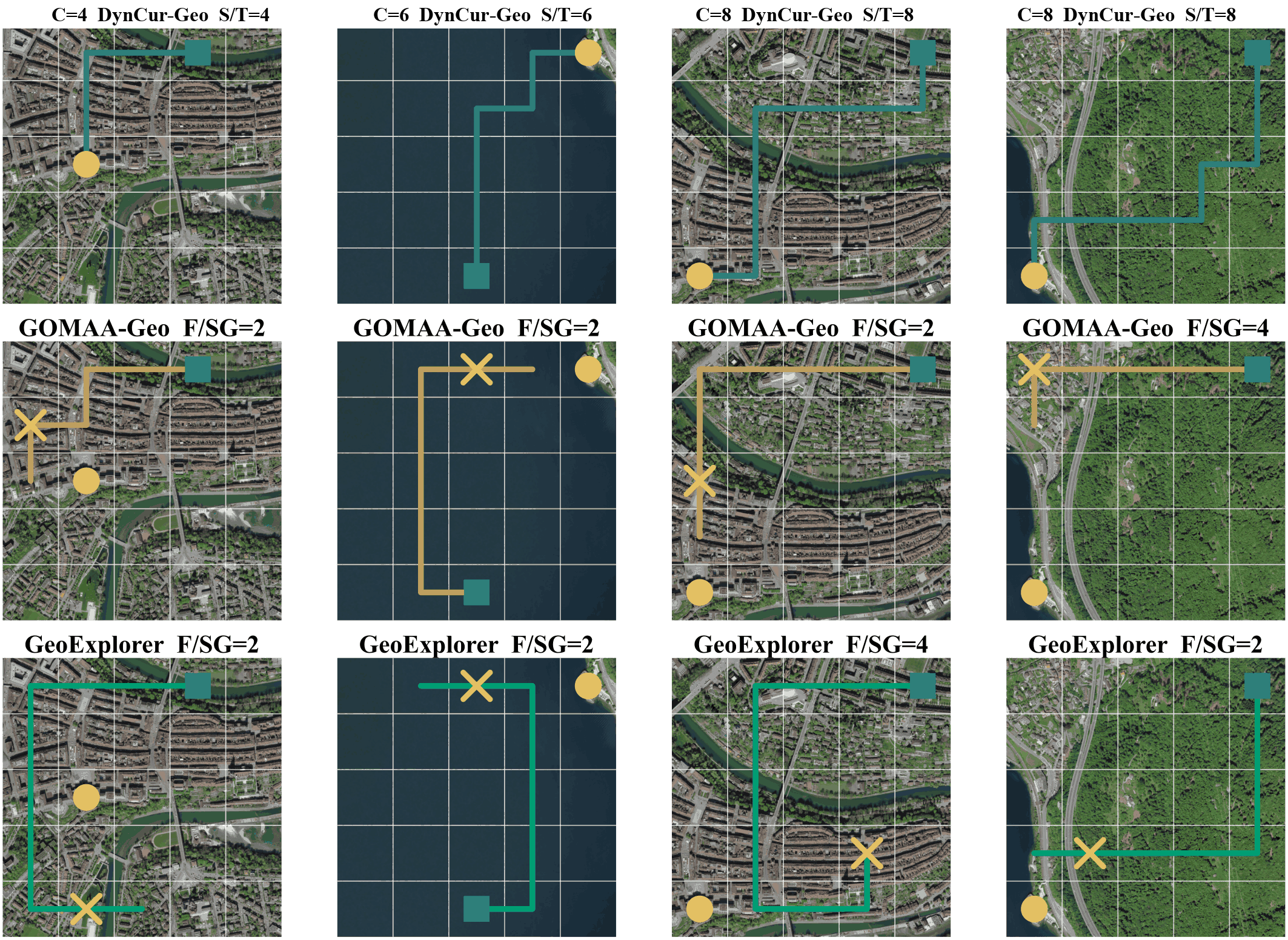}
\caption{Four same-task cases in which \methodname{} succeeds and both baselines fail, shown as matched task columns and method rows. The columns cover $C=4,6,8$ while giving each route panel sufficient space. Start, target, and budget are identical within a column; the route panels omit target-photo insets so that the paths remain unobstructed. $S/T$ gives success and path length; $F/SG$ gives failure and terminal distance.}
\label{fig:proposed_only}
\end{figure*}

Fig.~\ref{fig:proposed_only} displays representative proposed-only successes with matched task columns. The visible advantage follows two recurring geometries: avoiding a repeated local loop and retaining a promising target-neighborhood approach. Because every column fixes the task and target cue across policies, path differences reflect policy behavior under the same start--goal pair.

At $C=8$, the routes of \ourmethod{} show the clearest long-range pattern: exploration remains active while the target is distant, and the distance weight decreases near the final cells. The corresponding traces appear in the stepwise reward diagnostics below.

\begin{figure*}[t]
\centering
\includegraphics[width=\textwidth]{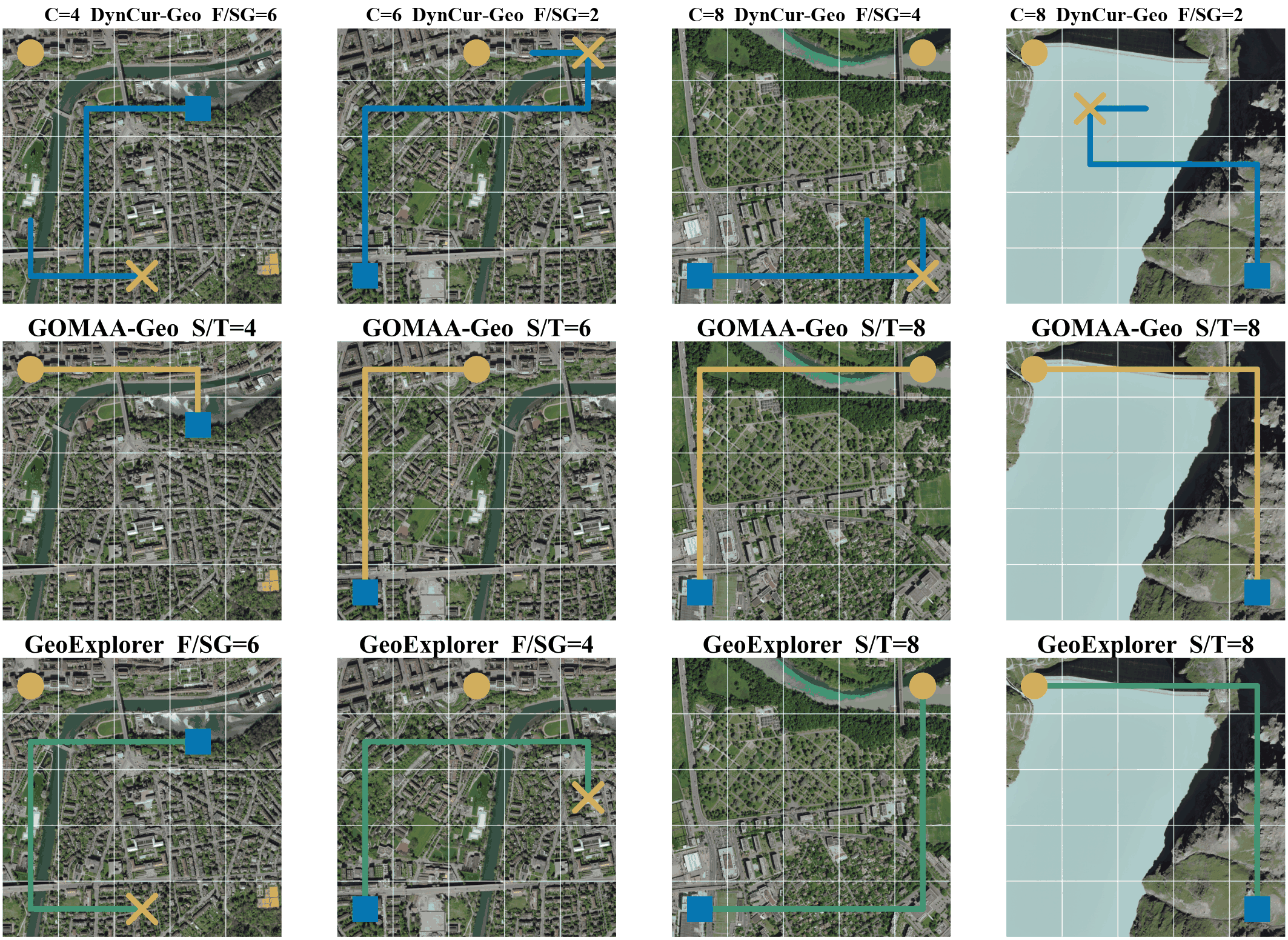}
\caption{Four matched cases in which \methodname{} fails and at least one baseline succeeds, again shown as matched task columns and method rows without target-photo insets. The layout and annotations match Fig.~\ref{fig:proposed_only}.}
\label{fig:counterexamples}
\end{figure*}

Fig.~\ref{fig:counterexamples} shows the complementary cases. Stronger long-distance averages can coexist with individual scenes in which a baseline retains the correct local hypothesis while \ourmethod{} loops, overshoots, or exits the target neighborhood.

These contrasts separate global progress from local target confirmation. A baseline may approach the correct neighborhood while matching a visually plausible distractor, whereas a successful route must preserve the target-conditioned hypothesis through the final transition.

The paired pattern frequencies show the same distance dependence. Complete three-policy failure falls from 41.3\% at $C=4$ to 9.7\% at $C=6$ and zero at $C=8$, while all-three success rises from 3.7\% to 18.0\% and 35.0\%. Longer routes provide more visible progress transitions, whereas short routes concentrate the decision around target-cell confirmation.

\section{Stepwise Reward Diagnostics}
\label{sec:reward_mechanism}

The training reward is
\begin{equation}
r_t=r_{\mathrm{ex},t}+\lambda_t r_{\mathrm{in},t}+r_{\Phi,t},
\end{equation}
where the linear distance-adjustment function is
\begin{align}
\rho_t &=
\operatorname{clip}\!\left(
\frac{d_1(p_t,p_g)}{\max(d_1(p_0,p_g),1)},0,1
\right)\\
\lambda_t &= \lambda_{\min}+(1-\lambda_{\min})\rho_t
\qquad \lambda_{\min}=0.405.
\end{align}

The extrinsic term gives the terminal hit a positive reward, gives an accepted progress transition a smaller positive reward, and penalizes invalid stays, revisits, and non-progress moves. In the recorded implementation, this accepted-progress branch uses the configured progress comparator, with the final checkpoint using the squared Euclidean cell comparator for that branch; Manhattan grid distance is still used for the displayed distance traces, for $\lambda_t$, and for the PBRS potential. The running best-progress state is initialized from the start--goal distance and, in the reported configuration, is committed after a reward-positive transition rather than after every displayed Manhattan decrease.

The intrinsic term is computed from the frozen environment-modeling transformer's next-state prediction error,
\begin{equation}
r_{\mathrm{in},t}=0.25\left[2\frac{\operatorname{MSE}(\hat h_{t+1},h_{t+1})-0.8}{0.1}-1\right],
\end{equation}
where $\hat h_{t+1}$ and $h_{t+1}$ are predicted and target hidden representations. The shaping term uses the following potential-based form.
\begin{equation}
\begin{gathered}
r_{\Phi,t}=\beta\!\left(\gamma\Phi(s_{t+1})-\Phi(s_t)\right)\\[-0.35ex]
\Phi(s_t)=-\frac{d_1(p_t,p_g)}{2K-2},
\end{gathered}
\end{equation}
with $\beta=0.10$, $\gamma=0.93$, and grid width $K=5$ for the standard setting. Target distance and reward terms are used for training and for the diagnostic plots. At test time, action selection uses the policy state and the legal-action mask. Recorded trajectories are replayed with their task targets to recover the reward decomposition along each sequence.

\begin{figure*}[t]
\centering
\includegraphics[width=\textwidth]{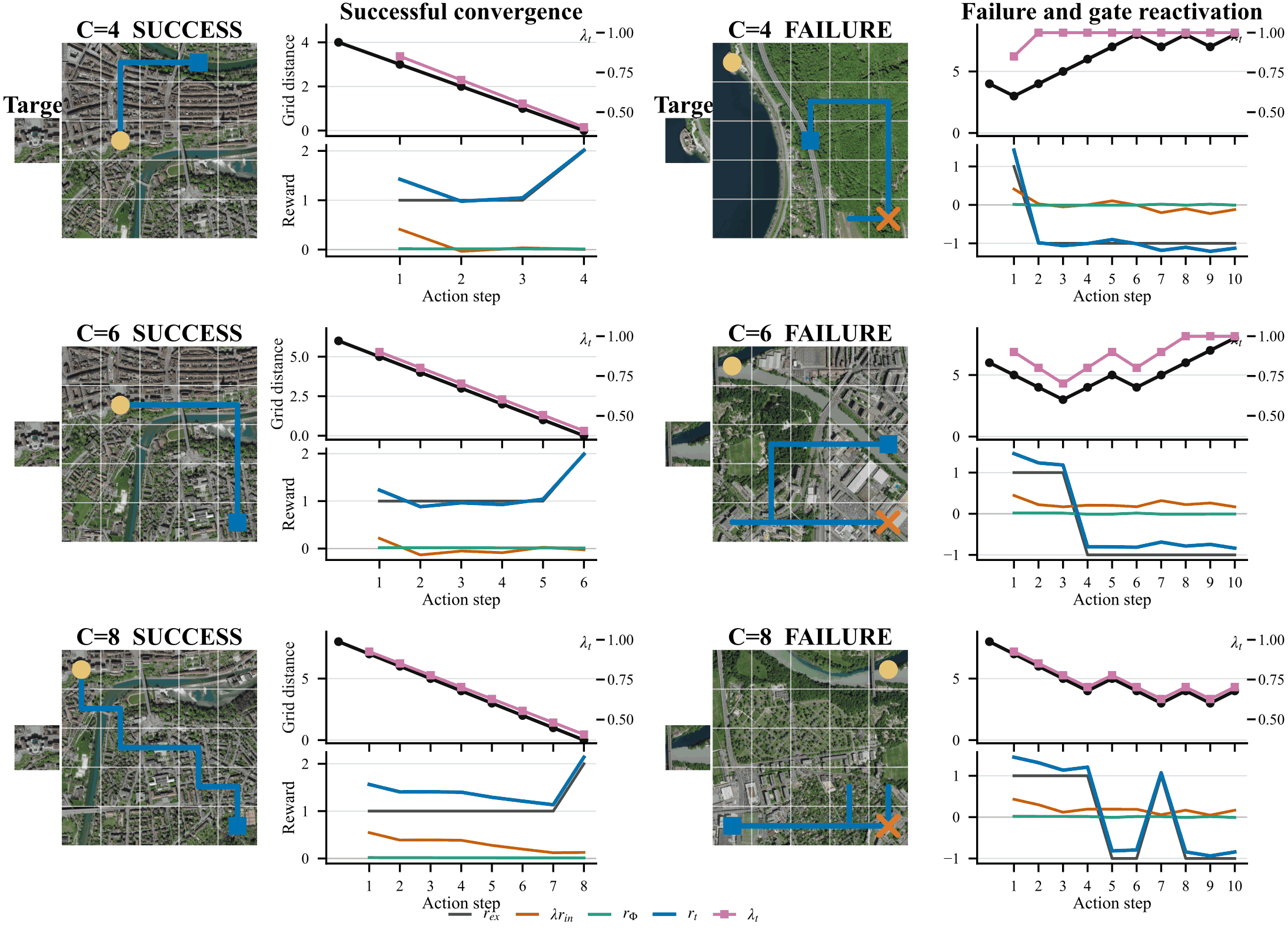}
\caption{Matched-distance stepwise diagnostics. Each row compares a success (left) with a failure (right) at the same $C$. Route panels include the exact target cue. Curves show remaining distance, the distance gate, and realized reward components. Successful routes contract the gate toward 0.405; failures show that regress actions reactivate the gate.}
\label{fig:joint_traces}
\end{figure*}

\begin{figure*}[p]
\centering
\includegraphics[width=0.99\textwidth]{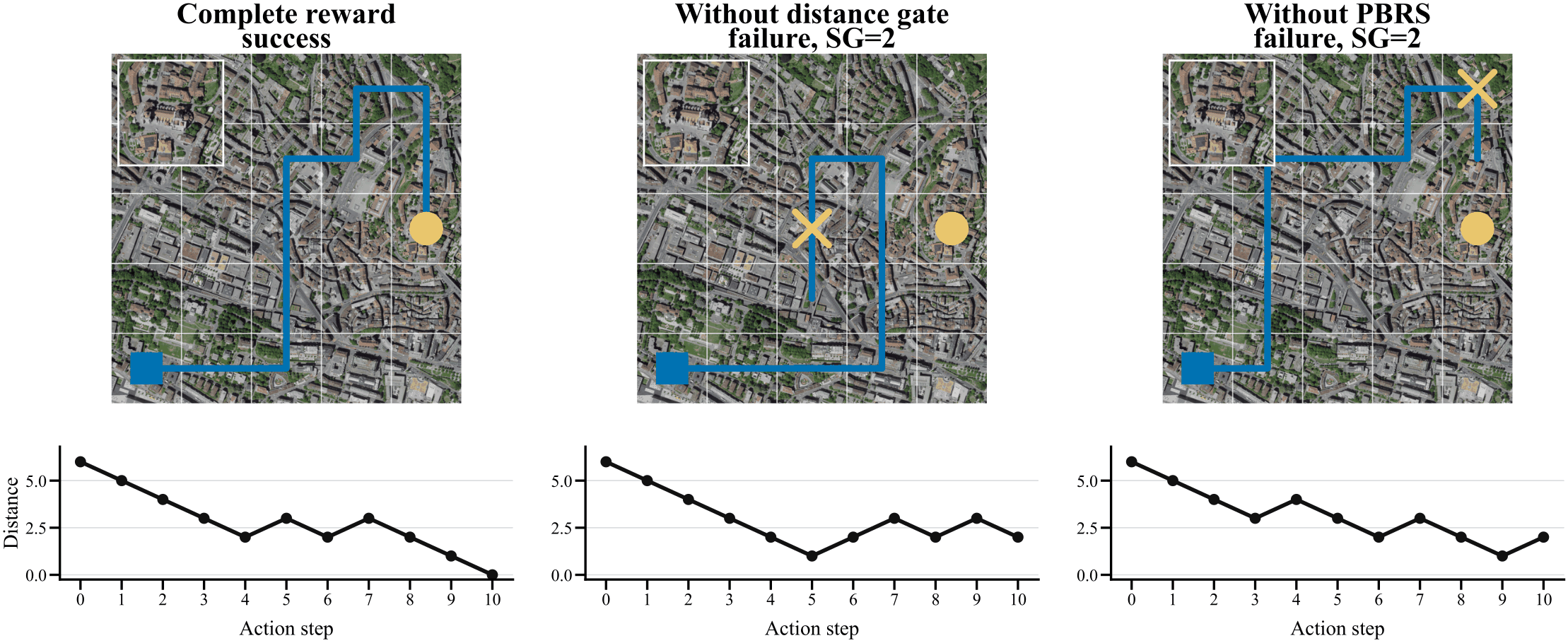}
\caption{Controlled route comparison on one fixed $C=6$ task. The complete-reward policy reaches the target; the separately trained policies without the distance gate or without PBRS both finish at SG $=2$. Only one training factor changes in each comparison.}
\label{fig:single_factor}
\end{figure*}

Fig.~\ref{fig:joint_traces} compares successes and failures at matched initial distances. In the three shortest-path successes, target distance decreases monotonically and the distance weight contracts on every transition. For $C=8$, it falls from 0.9256 after the first action to 0.405 at the target. The mean absolute raw intrinsic reward is 0.4323, while the mean absolute gated contribution is 0.3038. Across seven full-reward cases (six displayed traces plus the complete-reward control in Fig.~\ref{fig:single_factor}), the realized gated-to-raw magnitude ratio ranges from 0.405 to 1.0 and averages 0.7645.

\begin{figure*}[!t]
\centering
\includegraphics[width=0.94\textwidth]{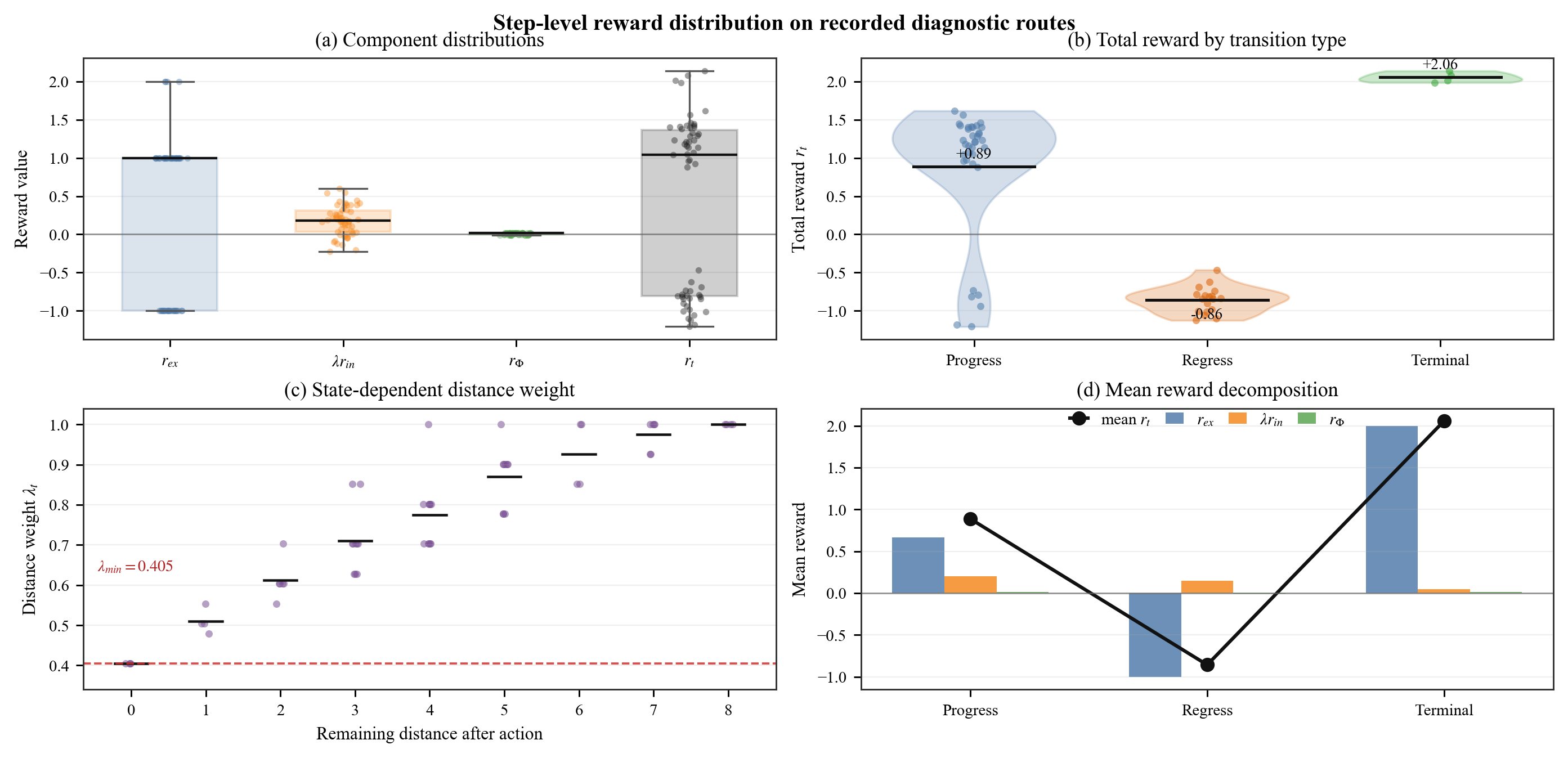}
\caption{Step-level reward distribution on the seven recorded full-reward diagnostic routes used for the mechanism plots. The summary covers 58 transitions and groups them by whether the executed action reduces target distance, increases target distance, or reaches the terminal target cell. The plot decomposes each transition into external reward, gated intrinsic contribution, PBRS shaping, and total reward; it is a diagnostic decomposition rather than a full-route-bank estimate.}
\label{fig:reward_distribution}
\end{figure*}

The three reward terms have different roles. External reward gives the dominant sign for target progress and the terminal hit. The intrinsic term is scene- and transition-dependent: it can be positive or negative after normalization and is not itself a target-direction signal. Potential shaping is small but directionally consistent. Over the same seven recorded full-reward cases, it averages (+0.0154) on 40 progress transitions and (-0.0076) on 18 regress transitions.

Fig.~\ref{fig:reward_distribution} shows the same decomposition after pooling transitions by realized movement type. Progress and terminal steps keep positive mean total reward, whereas regress steps remain negative despite a larger average distance weight. The mean distance weight is 0.7369 on progress transitions, 0.8995 on regress transitions, and 0.4050 at terminal hits. Thus, moving away from the target can reopen exploration pressure, but the total reward sign still favors target-directed motion.

The failures show where the same mechanism can lose the endpoint. In the $C=6$ failure, distance first falls from six to three, then oscillates and rises to eight; the distance weight falls to 0.7025 and returns to one as the route diverges. The $C=4$ failure makes one useful first move and then spends most of the budget increasing distance despite the fixed target cue. The $C=8$ failure preserves several correct transitions before entering a local oscillation. These rows separate early target-hypothesis selection at short range from target-hypothesis retention after a longer approach.

When a route moves away from the target, the distance weight increases again. This can help escape an incorrect local hypothesis, but it can also sustain motion toward a visually salient alternative. Mean total reward remains positive on progress transitions and negative on regress transitions, so the target-progress sign is preserved even when the trajectory later diverges. At the same action index, success and failure can carry different distance weights because their remaining distances differ.

The six traces quantify this split. The successful $C=4,6,8$ routes have mean distance weights 0.628, 0.653, and 0.665, and retain 0.800, 0.729, and 0.703 of the absolute raw intrinsic magnitude. The corresponding failure means are 0.985, 0.881, and 0.740, because regress motion repeatedly raises the weight. Mean total reward on regress actions remains negative in all three failures ($-1.027$, $-0.777$, and $-0.833$). These values describe the displayed diagnostics; the next subsection uses the full bank.

Fig.~\ref{fig:single_factor} compares independently trained controls on one fixed task. All three policies initially reduce target distance. The complete-reward policy survives several corrections and reaches the target on the tenth action. Without the distance weight, the route reaches distance one at step five but oscillates and finishes at distance two. Without PBRS, the route reaches distance one at step nine and then moves away. This case isolates one training factor at a time, while the aggregate ablation in the main paper reports the corresponding average effect.

\subsection{Population-Level Route Dynamics}

\begin{figure*}[p]
\centering
\includegraphics[width=0.99\textwidth]{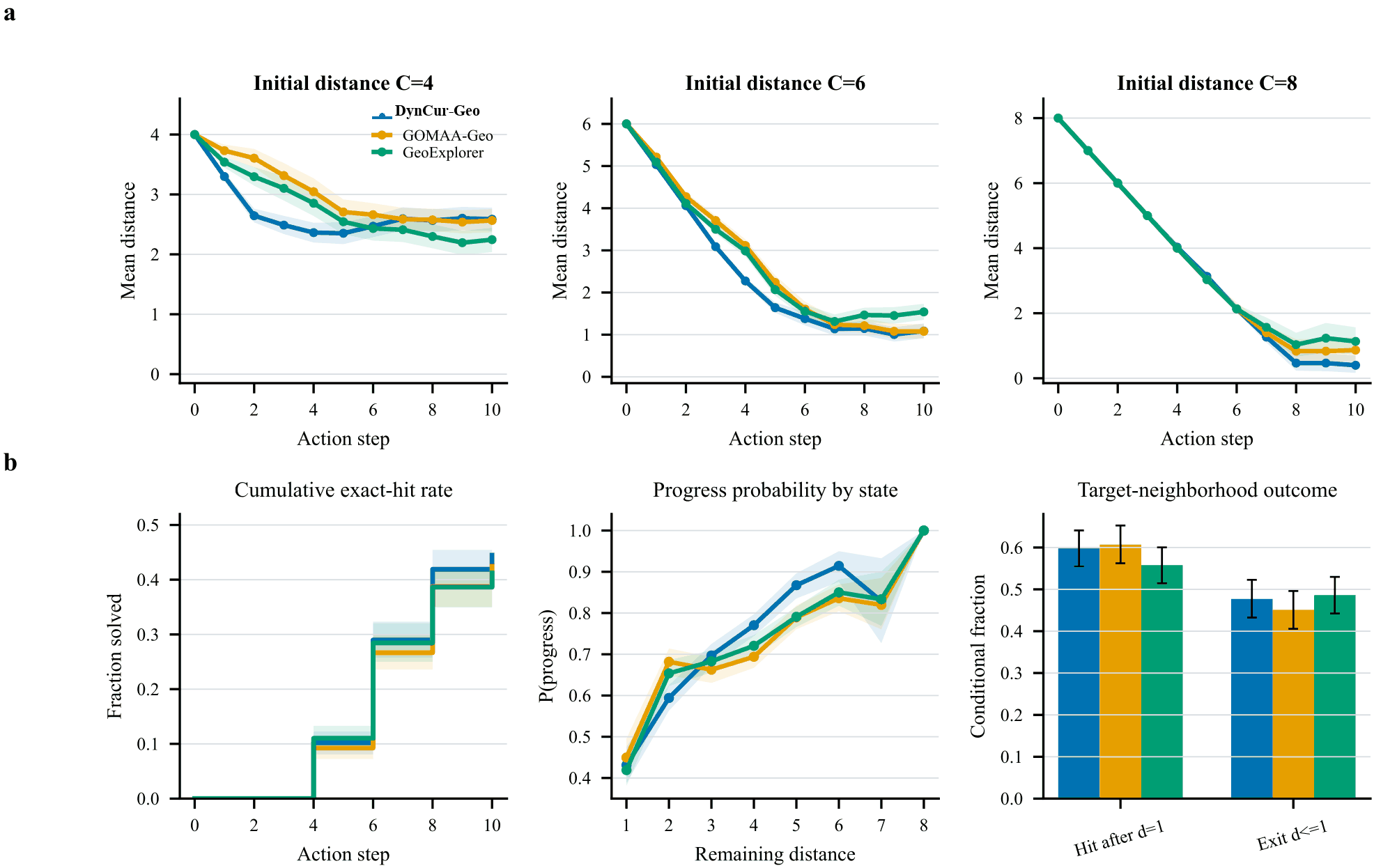}
\caption{Population dynamics over all 2,205 recorded routes. Bands and error bars are 95\% percentile intervals from 2,000 deterministic bootstrap resamples clustered by aerial image. Top: mean remaining distance by initial distance. Bottom: cumulative exact-hit rate, progress probability conditioned on current remaining distance, and outcomes after first reaching distance one. Curves summarize the full trajectory bank.}
\label{fig:population_dynamics}
\end{figure*}

Fig.~\ref{fig:population_dynamics} extends the six displayed traces to the complete bank. At $C=8$, \ourmethod{} follows an almost deterministic contraction through step seven and reaches a cumulative hit rate of 0.850 by step ten, compared with 0.683 for GOMAA-Geo and 0.600 for GeoExplorer. At $C=6$, its mean distance falls below both baselines by step four and remains lowest at termination. At $C=4$, early contraction stalls after step four and mean distance begins to rise. In this controlled bank, longer routes often provide several unambiguous progress transitions, whereas short routes expose target confirmation quickly.

The state-conditioned curve isolates the near-goal bottleneck. For \ourmethod{}, the probability of a progress action is 1.000 at remaining distance eight, 0.914 at six, 0.770 at four, and 0.431 at distance one; the complementary regress probability at distance one is 0.569. Among 547 routes that reach distance one, 59.8\% subsequently hit the target, while 47.7\% leave the target neighborhood at least once. The two events can both occur because a route may exit and later recover. The image-cluster intervals summarize sampling variation within the trajectory bank.

\section{Cross-Domain Targets and Enlarged Search Regions}
\label{sec:cross_domain}

The shared Swiss aerial bank isolates route geometry under one target form. This section reports additional results in which the target modality, search image domain, grid size, or action horizon changes while the evaluator remains target conditioned.

\subsection{Target/Search Evaluation Contracts}

All cross-domain experiments in this section use the same evaluator contract. A target cue identifies the requested destination before rollout, and the policy searches the corresponding observation stream under a fixed action budget. The cue may be an aerial crop, a ground photograph, a text-grounded target representation, or a pre-disaster target image. These forms change the evidence used to specify the goal, but success is still counted only when the trajectory reaches the target grid cell.

The displayed target cue defines the task instance and is encoded through the target-conditioned input pipeline used by the evaluator. It is not refreshed as a new observation at every step. Each route image shows the realized greedy rollout after legal-action masking. Aggregate tables summarize success under a protocol, while route figures show the corresponding endpoint decisions, local drift, and domain-shift effects.

Aerial-to-aerial tasks mainly test spatial search once the target appearance is directly available. MM-GAG adds multimodal target specification. SwissViewMonuments ground-to-aerial tasks test cross-view endpoint confirmation. xBD pre/post tasks keep the target fixed while changing the searched image domain. Equal aggregate SR can correspond to different route-level behavior.

\begin{figure*}[!t]
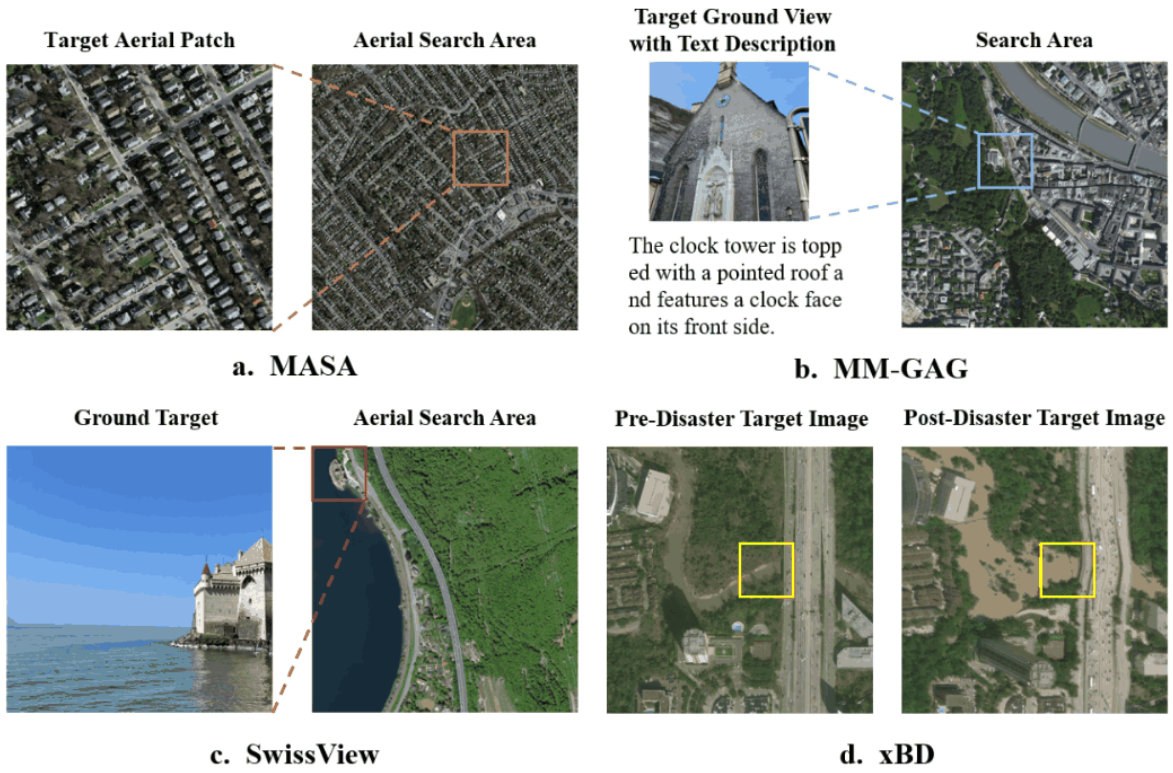

\centering
\sTenProtocolGraphic
\caption{Target/search protocol illustrations for aerial, MM-GAG, SwissView, and xBD inputs. The target cue is fixed before rollout, and the policy searches the corresponding observation stream.}
\label{fig:target_modalities}
\end{figure*}

Fig.~\ref{fig:target_modalities} specifies the target evidence, the searched image stream, and the input path used by the evaluator. The subsequent route figures then show the realized greedy rollout under that fixed contract. In Fig.~\ref{fig:xbd_routes}, the xBD pair is held constant except for the search domain. In Fig.~\ref{fig:mmgag_swiss_routes}, MM-GAG remains within one aerial family, whereas SwissViewMonuments requires ground-to-aerial endpoint confirmation. In Fig.~\ref{fig:long_range}, the stress test increases grid size and action horizon.

\subsection{xBD Pre/Post Same-Task Analysis}

The xBD reproduction uses the fixed pre/post evaluation task bank defined for the main experiments. Each setting contains episodes under $5\times5$, $B=12$, $C=4,\ldots,8$, five repeats, greedy actions, and task seed 20260516. The target is always a pre-disaster cell. The search image is pre-disaster in xBD-pre and post-disaster in xBD-disaster, so the task tests whether a policy trained for target-conditioned search remains stable when the search observation changes after damage.

\begin{figure*}[!t]
\centering
\includegraphics[width=0.96\textwidth]{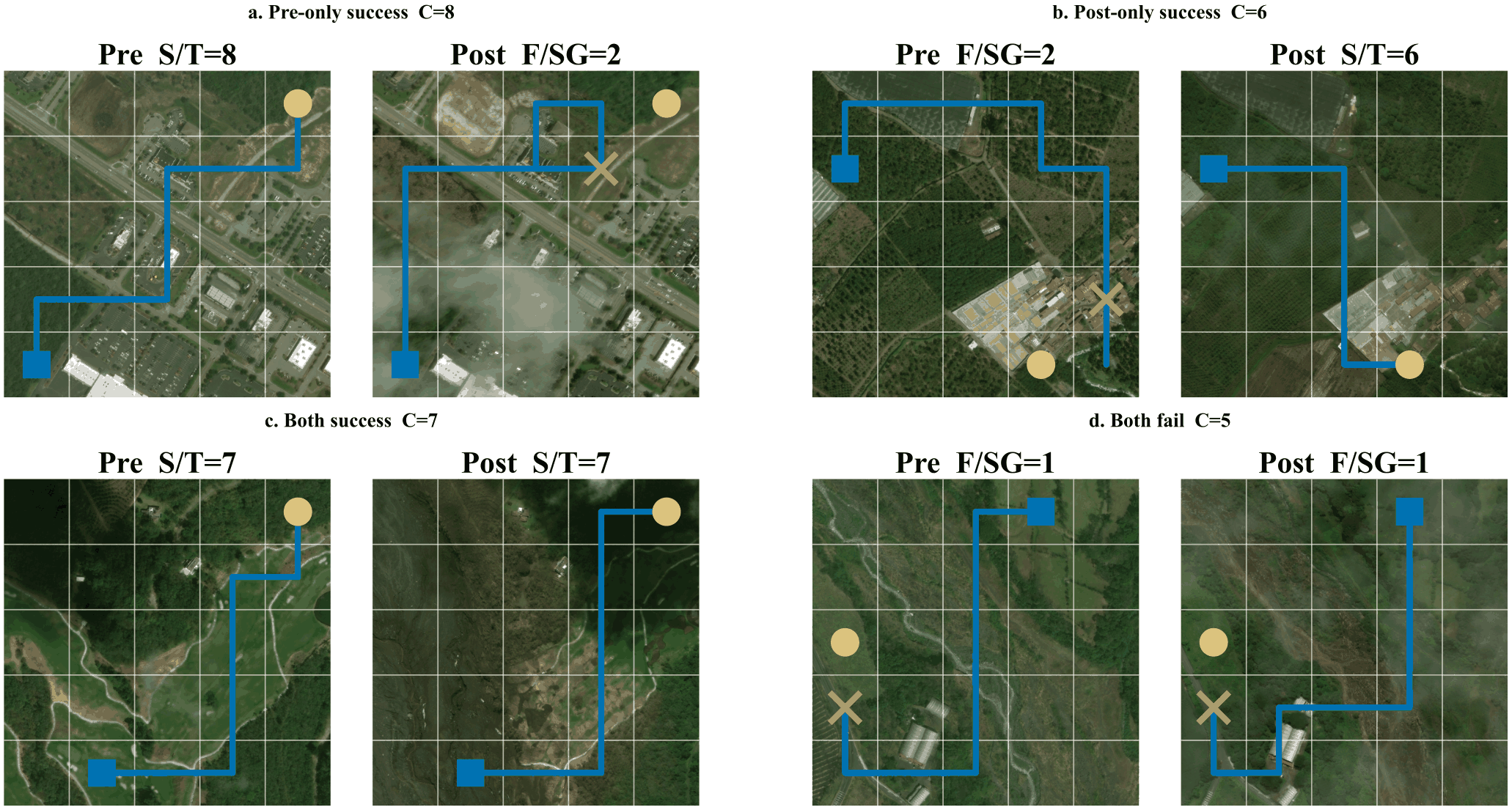}
\caption{Recorded xBD routes on four exactly matched tasks. Panels (a)--(d) show pre-only success, post-only success, success under both domains, and failure under both, respectively. Each panel pair fixes pair, start, pre-disaster target, $C$, repeat, and budget; only the search image changes from pre-disaster (left route) to post-disaster (right route). Target-cue thumbnails are omitted so the route panels remain large and unobstructed.}
\label{fig:xbd_routes}
\end{figure*}

Fig.~\ref{fig:target_modalities} specifies the target evidence and searched image stream. Fig.~\ref{fig:xbd_routes} fixes the task tuple and shows how the realized route changes when only the search domain changes. Aggregate xBD scores can remain nearly unchanged while individual paired tasks reverse outcome.

For all cross-domain cases, the evaluator uses the fixed target cue to define termination and metrics, keeping the panels comparable to the Swiss route bank.

The aggregate behavior is nearly invariant to the image shift. SR/SG are 0.5973/1.3018 before disaster and 0.5984/1.2813 after disaster. Across the 20,000 paired tasks, 9,353 succeed in both domains, 5,439 fail in both, 2,593 succeed only before disaster, and 2,615 succeed only after disaster. The discordant counts almost cancel at the aggregate level; Fig.~\ref{fig:xbd_routes} shows both directions of task-level reversal under the same start, target, distance, repeat, and budget.

Route dynamics are also stable. Successful path efficiency is 0.9304 before disaster and 0.9322 after disaster; revisit ratios are 0.1882 and 0.1888. Among episodes that reach distance one, 73.16\% and 73.32\% subsequently hit the exact target, while 34.65\% and 34.75\% leave the target neighborhood at least once. Under this reproduction split, appearance change modifies individual decisions without increasing aggregate drift.

\subsection{Aerial and Ground-View Target Routes}

\begin{figure*}[!t]
\centering
\includegraphics[width=0.84\textwidth]{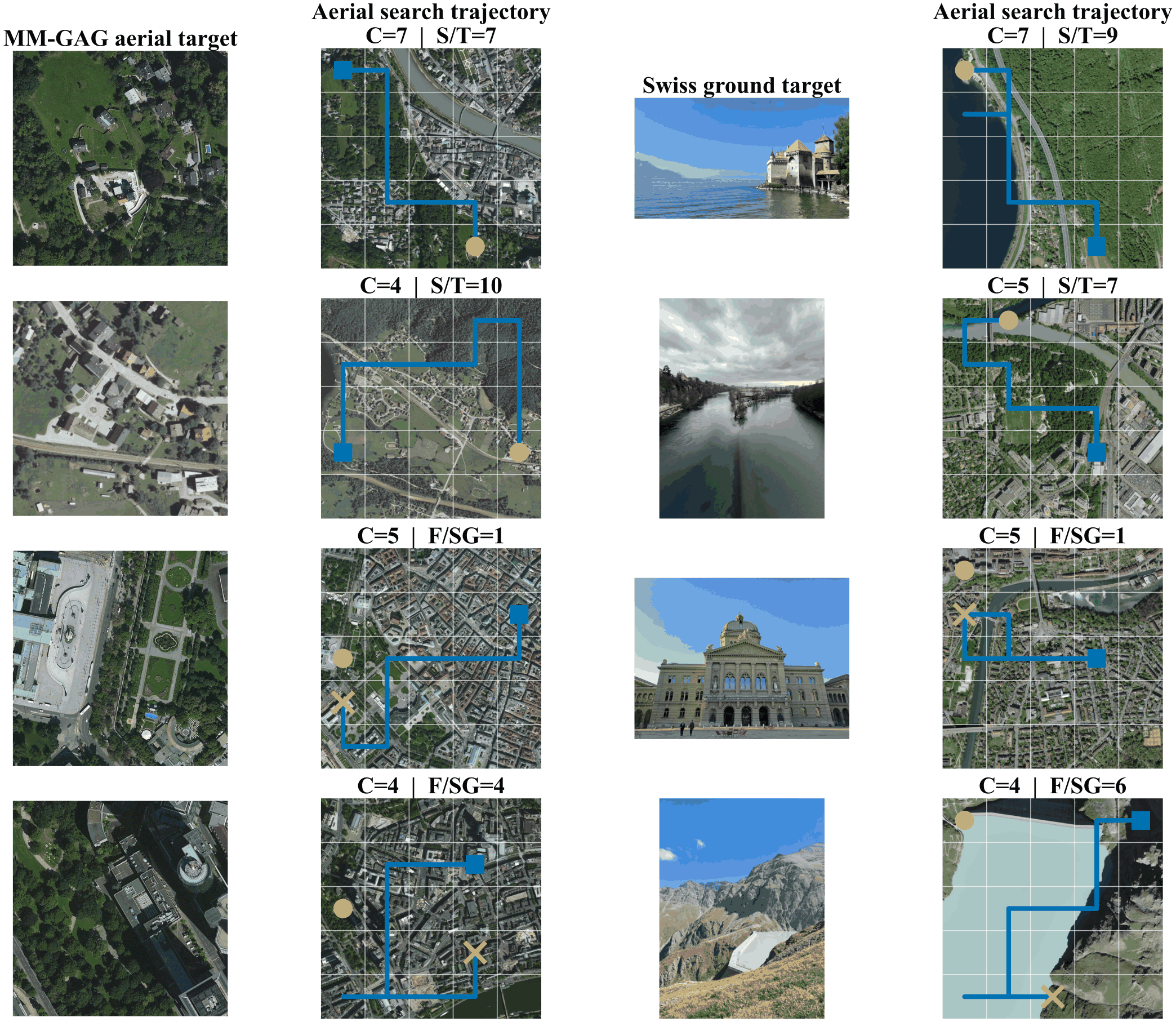}
\caption{Recorded cross-modal routes. Each row pairs an MM-GAG aerial target and route (left) with a SwissViewMonuments ground-photo target and aerial route (right). Cases cover long- and short-distance successes, near misses, and larger terminal errors. Swiss targets span castle, waterfront, civic-building, and dam/mountain scenes. The MM-GAG evaluator uses all 47 indexed samples; the Swiss rows use the controlled ground-to-aerial route set described in this supplement.}
\label{fig:mmgag_swiss_routes}
\end{figure*}

Fig.~\ref{fig:mmgag_swiss_routes} separates two target-confirmation regimes. In MM-GAG, the aerial cue and search image share a view family, so the policy must retain the target hypothesis across a larger stitched scene. The MM-GAG evaluator uses new target-conditioned tasks on the indexed image set rather than a held-out-image split. In SwissViewMonuments, the target is a ground photograph and the search image is aerial; the displayed rows span castle, waterfront, civic-building, and dam/mountain landmarks. This setting adds cross-view endpoint confirmation after global progress.

Table~\ref{tab:cross_domain_dynamics} shows a sharper endpoint problem for the Swiss ground-target setting. Its revisit ratio is 0.2176 versus 0.1732 for MM-GAG aerial targets, and 46.40\% of episodes that reach distance one later exit the neighborhood versus 36.24\%. Because the rows differ in both dataset and target modality, the comparison is descriptive rather than a controlled modality ablation.

\subsection{Long-Range Stress Test}

The aggregate SR of \ourmethod{} is 0.746, 0.748, and 0.204 on $8\times8$, $10\times10$, and $25\times25$, respectively, compared with 0.679, 0.629, and 0.180 for GOMAA-Geo. The largest grid remains difficult for all methods. On $25\times25$, SR varies strongly by distance stratum, rising from 0.02 at $C=12$ to 0.58 at $C=48$ for \ourmethod{}. The comparison is made under the same grid, budget, and distance setting for each row.

\begin{table*}[!t]
\centering
{\suppwidetable
\begin{tabular}{@{}>{\raggedright\arraybackslash}m{0.22\textwidth}*{8}{>{\centering\arraybackslash}m{0.078\textwidth}}@{}}
\toprule
Recorded evaluator & Episodes & SR & SG & Succ. eff. & Revisit & Progress & Hit after $d=1$ & Exit after $d=1$ \\
\midrule
xBD pre, aerial target & 20,000 & 0.5973 & 1.3018 & 0.9304 & 0.1882 & 0.7585 & 0.7316 & 0.3465 \\
xBD disaster, pre-disaster target & 20,000 & 0.5984 & 1.2813 & 0.9322 & 0.1888 & 0.7599 & 0.7332 & 0.3475 \\
MM-GAG, aerial target & 1,175 & 0.6357 & 1.1319 & 0.9278 & 0.1732 & 0.7720 & 0.7477 & 0.3624 \\
SwissMon ground, max80 & 1,380 & 0.5420 & 1.4457 & 0.8714 & 0.2176 & 0.7124 & 0.6997 & 0.4640 \\
\bottomrule
\end{tabular}
\par}
\caption{Route-level descriptors recomputed from every saved trajectory in four evaluators. Succ. eff. is shortest distance divided by path length on successful episodes. The last two columns condition on first reaching distance one. The rows have different dataset scopes and are diagnostic rather than a controlled modality ablation.}
\label{tab:cross_domain_dynamics}
\end{table*}

\begin{figure*}[!t]
\centering
\includegraphics[width=0.90\textwidth]{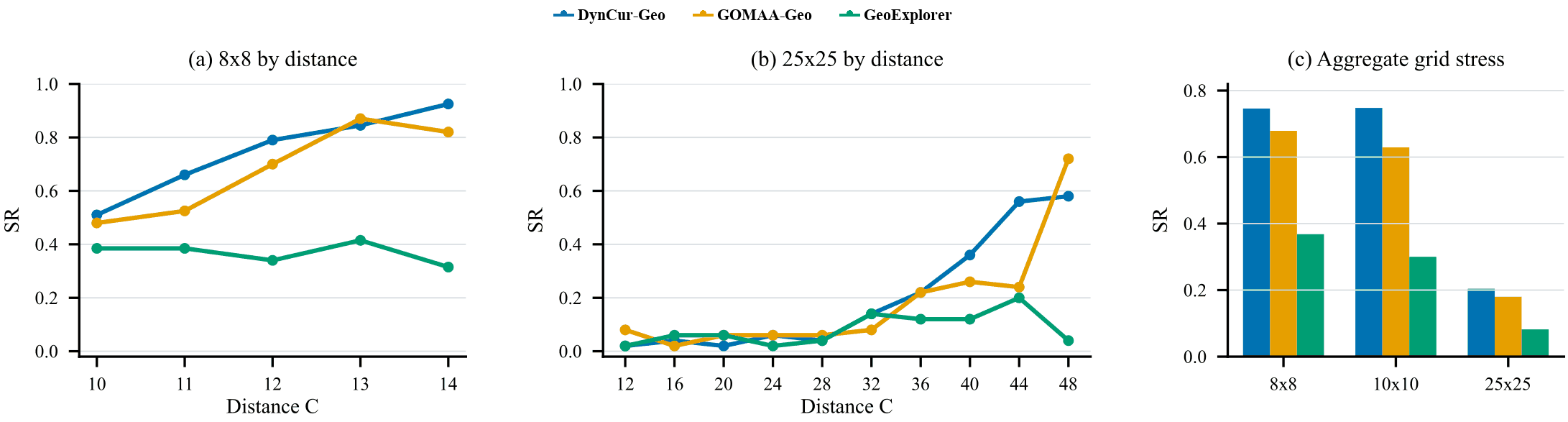}
\caption{Formal MASA enlarged-grid results. Panels (a)--(b) report per-distance SR for $8\times8$ ($B=24$, $C=10,\ldots,14$) and $25\times25$ ($B=60$, $C=12,16,\ldots,48$); panel (c) reports aggregate SR for $8\times8$, $10\times10$, and $25\times25$, where $10\times10$ uses $B=32$ and $C=14,\ldots,18$.}
\label{fig:long_range}
\end{figure*}

\section{Additional Quantitative Evidence}

This section reports additional numerical studies associated with the main-paper protocol. The tables cover training-pool selection, distance-stratified results for the unified scenarios, development hyperparameters, and enlarged-grid stress tests. They use the same evaluation definitions unless a separate grid size, budget, or distance range is stated.

\subsection{Training-Pool Selection}

The training image pool is not a fixed route list. For each sampled image, training episodes are generated online by resampling a start cell, a goal cell, and an initial-distance constraint. Dynamic sampling yields many target-conditioned navigation episodes from the same image pool, while evaluation task banks remain fixed. Table~\ref{tab:training_pool_selection} reports the controlled training-pool study. Average SR/SG values are macro-averaged over nine evaluation settings: MASA/A, MM-GAG/A/G/T, SwissView100/A, SwissViewMonuments/A/G, xBD-pre/A, and xBD-disaster/A.

\begin{table}[!t]
\centering
{\suppnarrowtable
\begin{tabularx}{\columnwidth}{@{}>{\raggedright\arraybackslash}X>{\centering\arraybackslash}p{0.13\columnwidth}>{\centering\arraybackslash}p{0.18\columnwidth}>{\centering\arraybackslash}p{0.18\columnwidth}@{}}
\toprule
Training pool & Images & Avg. SR & Avg. SG \\
\midrule
MASA+MM-GAG & 184 & 0.6033 & 1.2543 \\
MASA+SwissView100 & 237 & 0.5684 & 1.4191 \\
MM-GAG+SwissView100 & 147 & 0.5497 & 1.7869 \\
MASA+MM-GAG+SwissView100 & 284 & 0.5232 & 1.7952 \\
SwissView100 & 100 & 0.5103 & 2.0490 \\
MASA & 137 & 0.4443 & 2.1380 \\
MM-GAG & 47 & 0.4328 & 2.0731 \\
\bottomrule
\end{tabularx}
\par}
\caption{Training-pool selection under the unified evaluator. The final pool uses MASA+MM-GAG: 137 MASA images and 47 MM-GAG images. The comparison uses dynamically sampled navigation episodes from each image pool.}
\label{tab:training_pool_selection}
\end{table}

The best aggregate result comes from the 184-image MASA+MM-GAG pool. MASA provides the base aerial search distribution, while MM-GAG adds target-modality diversity. Adding SwissView100 increases the image count but reduces the macro-average. The final pool uses the smaller MASA+MM-GAG combination.

\subsection{Distance-Stratified Main Scenarios}

The main paper reports the unified $B=12$ comparison as scenario-level averages. Table~\ref{tab:distance_stratified_main} adds the distance decomposition for the same rows of \ourmethod{}. Short-distance cases are not always easier: at $C=4$ and $C=5$, many failures are endpoint-confirmation errors near the target, while longer settings provide more room for repeated progress transitions before the final cell is reached.

\begin{table}[!t]
\centering
{\small\setlength{\tabcolsep}{1.2pt}\renewcommand{\arraystretch}{1.00}
\makebox[\columnwidth][c]{\textbf{(a) Success rate by scenario and initial distance}}\par\vspace{0.35mm}
\begin{tabular*}{\columnwidth}{@{\extracolsep{\fill}}lrrrrrr@{}}
\toprule
Scenario & Avg. & $C=4$ & $C=5$ & $C=6$ & $C=7$ & $C=8$ \\
\midrule
MASA/A & 0.6200 & 0.2400 & 0.4000 & 0.6400 & 0.8800 & 0.9400 \\
MM-GAG/A & 0.6357 & 0.2979 & 0.3915 & 0.6894 & 0.8766 & 0.9234 \\
xBD-pre/A & 0.5973 & 0.3033 & 0.4065 & 0.6230 & 0.8115 & 0.8423 \\
xBD-dis./A & 0.5984 & 0.2923 & 0.4085 & 0.6290 & 0.8180 & 0.8443 \\
\bottomrule
\end{tabular*}
\par\vspace{1.0mm}
\makebox[\columnwidth][c]{\textbf{(b) Final grid distance by scenario and initial distance}}\par\vspace{0.35mm}
\begin{tabular*}{\columnwidth}{@{\extracolsep{\fill}}lrrrrrr@{}}
\toprule
Scenario & Avg. & $C=4$ & $C=5$ & $C=6$ & $C=7$ & $C=8$ \\
\midrule
MASA/A & 1.1360 & 2.6800 & 1.5200 & 1.0000 & 0.2400 & 0.2400 \\
MM-GAG/A & 1.1319 & 2.5021 & 1.6894 & 0.8681 & 0.3617 & 0.2383 \\
xBD-pre/A & 1.3018 & 2.4470 & 1.8755 & 1.1565 & 0.5125 & 0.5175 \\
xBD-dis./A & 1.2813 & 2.4475 & 1.8685 & 1.0945 & 0.4960 & 0.5000 \\
\bottomrule
\end{tabular*}
\par}
\caption{Distance-stratified \methodname{} results for the unified $5\times5$, $B=12$, $C=4,\ldots,8$ evaluator. Panel (a) reports SR and panel (b) reports SG. The same fixed task-bank protocol is used as in the unified main comparison.}
\label{tab:distance_stratified_main}
\end{table}

\subsection{Development Hyperparameter Checks}

Table~\ref{tab:development_hyperparams} reports development checks for entropy, the distance lower bound, PBRS, training budget, and validation-distance selection. These rows are not separate baselines.

\begin{table}[!t]
\centering
{\small
\setlength{\tabcolsep}{2.0pt}
\renewcommand{\arraystretch}{0.98}
\textbf{(a) Entropy coefficient \(c_e\)}
\begin{tabularx}{\columnwidth}{@{}l*{4}{Y}@{}}
\toprule
 & 0.0025 & 0.0050 & 0.0075 & 0.0100 \\
\midrule
Avg. SR & 0.5497 & \textbf{0.5765} & 0.4344 & 0.4729 \\
Avg. SG & 1.4166 & \textbf{1.2885} & 2.0582 & 1.7864 \\
\bottomrule
\end{tabularx}
\par\vspace{0.6mm}
\textbf{(b) Distance lower bound \(\lambda_{\min}\)}
\begin{tabularx}{\columnwidth}{@{}l*{5}{Y}@{}}
\toprule
 & 0.250 & 0.350 & 0.405 & 0.500 & 0.650 \\
\midrule
Avg. SR & 0.5009 & 0.4861 & \textbf{0.5765} & 0.4987 & 0.5089 \\
Avg. SG & 1.7052 & 1.6788 & \textbf{1.2885} & 1.6200 & 1.6526 \\
\bottomrule
\end{tabularx}
\par\vspace{0.6mm}
\textbf{(c) PBRS coefficient \(\beta\)}
\begin{tabularx}{\columnwidth}{@{}l*{5}{Y}@{}}
\toprule
 & 0.00 & 0.05 & 0.10 & 0.15 & 0.20 \\
\midrule
Avg. SR & 0.5535 & 0.5134 & \textbf{0.5765} & 0.4933 & 0.5307 \\
Avg. SG & 1.3290 & 1.5409 & \textbf{1.2885} & 1.7120 & 1.4880 \\
\bottomrule
\end{tabularx}
\par\vspace{0.6mm}
\textbf{(d) Training and checkpoint selection}
\begin{tabularx}{\columnwidth}{@{}lYYY@{}}
\toprule
Budget & 240k & 480k & 720k \\
\midrule
Avg. SR & 0.4923 & \textbf{0.5765} & 0.5673 \\
Avg. SG & 1.6221 & \textbf{1.2885} & 1.2851 \\
\midrule
Val. \(C\) & 4--8 & 6--8 & 7--8 \\
\midrule
Avg. SR & 0.4282 & 0.5334 & \textbf{0.5765} \\
Avg. SG & 2.1324 & 1.4900 & \textbf{1.2885} \\
\bottomrule
\end{tabularx}
\par}
\caption{Additional development checks under the standard $5\times5$, $B=10$, $C=4,\ldots,8$ evaluator. The selected setting uses $c_e=0.005$, $\lambda_{\min}=0.405$, $\beta=0.10$, and 480k PPO steps.}
\label{tab:development_hyperparams}
\end{table}

\subsection{Detailed Enlarged-Grid Results}

Fig.~\ref{fig:long_range} summarizes the enlarged-grid stress tests. Tables~\ref{tab:long_range_detail} and~\ref{tab:long_range_25} give the numerical decomposition. The $8\times8$ and $10\times10$ settings show gains over both baselines across most distance strata. The $25\times25$ setting remains difficult for all methods, but \ourmethod{} has the best aggregate SR/SG and higher SR at several larger distance bins.

\begin{table}[!ht]
\centering
{\small\setlength{\tabcolsep}{0pt}\renewcommand{\arraystretch}{1.00}
\makebox[\columnwidth][c]{\textbf{(a) $8\times8$, $B=24$, MASA/A}}\\[-0.5mm]
\begin{tabular*}{\columnwidth}{@{\extracolsep{\fill}}lcrrrrr@{}}
\toprule
Method & SR/SG & $10$ & $11$ & $12$ & $13$ & $14$ \\
\midrule
GOMAA-Geo & 0.679/1.345 & 0.480 & 0.525 & 0.700 & 0.870 & 0.820 \\
GeoExplorer & 0.368/3.046 & 0.385 & 0.385 & 0.340 & 0.415 & 0.315 \\
\methodname{} & 0.746/0.859 & 0.510 & 0.660 & 0.790 & 0.845 & 0.925 \\
\bottomrule
\end{tabular*}
\par\vspace{1.0mm}
\makebox[\columnwidth][c]{\textbf{(b) $10\times10$, $B=32$, MASA/A}}\\[-0.5mm]
\begin{tabular*}{\columnwidth}{@{\extracolsep{\fill}}lcrrrrr@{}}
\toprule
Method & SR/SG & $14$ & $15$ & $16$ & $17$ & $18$ \\
\midrule
GOMAA-Geo & 0.629/1.801 & 0.535 & 0.475 & 0.625 & 0.750 & 0.760 \\
GeoExplorer & 0.300/4.677 & 0.335 & 0.385 & 0.335 & 0.220 & 0.225 \\
\methodname{} & 0.748/0.736 & 0.540 & 0.675 & 0.755 & 0.885 & 0.885 \\
\bottomrule
\end{tabular*}
\par}
\caption{Per-distance SR for the $8\times8$ and $10\times10$ enlarged-grid stress tests. Numeric headers give the initial distance $C$; the SR/SG column reports the aggregate pair for each grid.}
\label{tab:long_range_detail}
\end{table}

\begin{table}[!ht]
\centering
{\small\setlength{\tabcolsep}{0pt}\renewcommand{\arraystretch}{1.00}
\makebox[\columnwidth][c]{\textbf{(a) $25\times25$, $B=60$, $C=12,\ldots,28$}}\\[-0.5mm]
\begin{tabular*}{\columnwidth}{@{\extracolsep{\fill}}lcrrrrr@{}}
\toprule
Method & SR/SG & $12$ & $16$ & $20$ & $24$ & $28$ \\
\midrule
GOMAA-Geo & 0.180/13.508 & 0.080 & 0.020 & 0.060 & 0.060 & 0.060 \\
GeoExplorer & 0.082/15.584 & 0.020 & 0.060 & 0.060 & 0.020 & 0.040 \\
\methodname{} & 0.204/12.312 & 0.020 & 0.040 & 0.020 & 0.060 & 0.040 \\
\bottomrule
\end{tabular*}
\par\vspace{1.0mm}
\makebox[\columnwidth][c]{\textbf{(b) $25\times25$, $B=60$, $C=32,\ldots,48$}}\\[-0.5mm]
\begin{tabular*}{\columnwidth}{@{\extracolsep{\fill}}lrrrrr@{}}
\toprule
Method & $32$ & $36$ & $40$ & $44$ & $48$ \\
\midrule
GOMAA-Geo & 0.080 & 0.220 & 0.260 & 0.240 & 0.720 \\
GeoExplorer & 0.140 & 0.120 & 0.120 & 0.200 & 0.040 \\
\methodname{} & 0.140 & 0.220 & 0.360 & 0.560 & 0.580 \\
\bottomrule
\end{tabular*}
\par}
\caption{Per-distance SR for the $25\times25$ enlarged-grid stress test. Numeric headers give the initial distance $C$; panel (a) also lists aggregate SR/SG for each method, and panel (b) continues the larger distance strata.}
\label{tab:long_range_25}
\end{table}

\section{Discussion and Limitations}

\subsection{Supported Claims}

The route evidence supports three claims. Recorded successful routes have high efficiency, low revisit ratios, and few immediate reversals relative to their displayed target cues. The realized distance weight attenuates intrinsic reward as remaining distance falls and increases it again when a route diverges. Potential shaping changes sign with progress versus regress transitions. The complete-bank curves further show high progress probability when the target is distant and lower progress probability in the immediate target neighborhood.

The shared-bank SR values are close for the three methods, but the paired counts show different route geometries. \ourmethod{} contributes additional successes on long or loop-prone tasks, while baseline successes remain visible on several short-range endpoint cases.

\subsection{Why Failures Persist}

Distance gating regulates curiosity during search, but near-goal target confirmation remains the most sensitive stage. More than half of recorded failures enter the target neighborhood and then drift. Across all policies, 55.8--60.7\% of routes that first reach distance one subsequently hit the target, and 45.1--48.6\% leave that neighborhood at least once. A nonzero curiosity floor preserves adaptability near the goal, but the fixed action budget still requires exact final-cell resolution.

Two behavior patterns account for most remaining errors. Recurrent history reduces but does not eliminate repeated-cell behavior; the 133 revisit/ping-pong failures concentrate in this category. When a route moves away from the target, the intrinsic weight rises again. This can support recovery from an incorrect local hypothesis, but also places greater demand on the visual state representation.

\subsection{Evidence Boundaries}

The principal trajectory bank uses SwissViewMonuments aerial targets, $C\in\{4,6,8\}$, and one fixed task seed. The cross-domain section adds ground-target and disaster routes plus enlarged-grid aggregates with their own budgets and dataset scopes. The success and failure atlases cover behavior categories, while the aggregate tables provide the corresponding frequency summaries.

The taxonomy uses explicit thresholds. Its two dominant geometries are reaching distance one before moving away, and revisiting or immediately reversing. Image-level explanations are treated as qualitative interpretations rather than encoder-level attribution.

\section{Reproducibility Details}

The Swiss route panels are reconstructed from recorded sequences over $1500\times1500$ aerial images. Route views are uncropped; footers report the target cell and, when applicable, the terminal cell. Cross-domain route panels follow the same convention for xBD, MM-GAG, and Swiss ground-to-aerial cases.

Case selection follows deterministic route descriptors: four cases per success category, four per failure category, and matched tasks across contrast figures. The Swiss route set contains 118 displayed route-panel records, and the cross-domain route set adds 16 evaluator panels. Reward figures replay fixed routes with task targets for diagnostic decomposition, population curves use all 2,205 sequences with 2,000 fixed-seed aerial-image-cluster bootstraps, and enlarged-grid results are reported as aggregate and per-distance stress tests.

The experiments use existing datasets rather than introducing a new dataset. MASA, MM-GAG, SwissViewMonuments, and xBD follow the task definitions in the main paper. Fixed task banks specify image id, start cell, target cell, distance stratum, repeat index, grid size, and action budget; route images are rendered from saved trajectories after evaluation.

Route figures expose search geometry, endpoint drift, and same-task reversals. Aggregate tables summarize evaluator outputs across the corresponding route sets.

\subsection{Experimental Configuration}

Unless otherwise stated, the standard evaluator uses a $5\times5$ grid, budget $B=12$, distance strata $C=4,\ldots,8$, five fixed task repeats per distance, greedy decoding, legal-action masking, and task-bank seed 20260516. The xBD pre/post analysis uses the fixed pre/post task bank defined for the main experiments. The enlarged-grid stress tests use the protocols in Fig.~\ref{fig:long_range}: $8\times8$ with $B=24$ and $C=10,\ldots,14$, $10\times10$ with $B=32$ and $C=14,\ldots,18$, and $25\times25$ with $B=60$ and $C=12,16,\ldots,48$.

The final PPO configuration uses actor and critic learning rates of $10^{-4}$, discount factor $\gamma=0.93$, four PPO epochs per update, clipping parameter $0.2$, and learning-rate decay $0.9999$. The state-action transformer is loaded from its pretrained checkpoint and kept frozen during PPO. The reported training seed is 321. The 480k value is the PPO budget, and the reported model row uses the validation-selected checkpoint. The reward follows the main paper, including the distance-dependent intrinsic weight with floor $0.405$ and the potential-based shaping term.

Reported GPU experiments were run on a Linux x86\_64 workstation with four NVIDIA GeForce RTX 3080 GPUs, 20 GB memory per GPU. The main software stack was Python 3.11, PyTorch 2.0.1, CUDA 11.7 runtime packages, torchvision 0.15.2, Transformers 4.39.2, NumPy 1.25.2, pandas 2.0.3, SciPy 1.14.1, scikit-learn 1.4.0, and Matplotlib 3.8.2.

Training and evaluation randomness are separated. The reported checkpoint uses the fixed training seed above, while the evaluator uses task-bank seed 20260516 for start/target tuples. Route-atlas case selection is deterministic, and the population bands use a fixed 2,000-replicate aerial-image-cluster bootstrap; these intervals summarize variation within the saved route bank rather than across independently trained seeds.

\subsection{Implementation-Level Reproducibility}

Each inference episode is an ordered patch sequence initialized with the target embedding and the first observed search patch. After every selected action, the evaluator appends the next search-patch embedding. The frozen transformer receives this sequence together with visited patch indices and previous action tokens. The PPO actor and critic operate on the resulting hidden state, so the action head uses transformer features rather than raw patch tokens alone.

Before action selection, a legal-action mask removes outward moves at grid boundaries and renormalizes the remaining action probabilities. Formal evaluation uses greedy decoding and stops only after reaching the target patch or exhausting the budget. The policy does not receive target distance, reward values, success flags, or diagnostic categories at test time.

Training rewards remain outside the inference input. The extrinsic term rewards exact hits and target-progress transitions while penalizing repeats, non-progress, and invalid stays. The intrinsic term is the mean-squared next-state prediction error between predicted and target hidden representations, multiplied by the distance-dependent weight. The potential term uses the change in normalized distance potential between consecutive states.

\end{document}